\documentclass[sigconf, nonacm]{acmart}

\usepackage[utf8]{inputenc} 
\usepackage[T1]{fontenc}    
\usepackage{xr}
\usepackage{url}            
\usepackage{booktabs}       
\usepackage{amsfonts}       
\usepackage{nicefrac}       
\usepackage{microtype}      
\usepackage{times}          
\usepackage{helvet}         
\usepackage{courier}        
\usepackage{url}            
\usepackage{graphicx}       
\usepackage{wrapfig}
\usepackage{subcaption}
\usepackage{placeins}
\usepackage{bm}
\usepackage{float}
\usepackage[disable]{todonotes}

\renewcommand{\P}{\ensuremath{\mathcal{P}}}
\newcommand{\D}{\ensuremath{\mathcal{D}}}
\newcommand{\xv}{\ensuremath{\bm{x}}}
\newcommand{\lv}{\ensuremath{\bm{\lambda}}}
\newcommand{\fhx}{\ensuremath{\hat{f}(\bm{x})}}

\newcommand{\Al}{\ensuremath{\mathcal{A}_{\bm{\lambda}}}}
\newcommand{\AlD}{\ensuremath{\mathcal{A}_{\bm{\lambda}}(\mathcal{D})}}
\newcommand{\Rl}{\ensuremath{R(\bm{\lambda})}}
\newcommand{\RlP}{\ensuremath{R_{\P}(\bm{\lambda})}}
\newcommand{\Rlk}{\ensuremath{R_k(\bm{\lambda})}}
\newcommand{\reals}{\ensuremath{\mathbb{R}}}

\makeatletter
\newcommand*{\addFileDependency}[1]{
  \typeout{(#1)}
  \@addtofilelist{#1}
  \IfFileExists{#1}{}{\typeout{No file #1.}}
}
\makeatother

\begin{document}

\title{Meta-Learning for Symbolic Hyperparameter Defaults}

\author{Pieter Gijsbers}
\authornotemark[1]
\affiliation{
  \institution{University of Eindhoven}
  \city{Eindhoven}
  \country{Netherlands}
}

\author{Florian Pfisterer}
\affiliation{%
  \institution{Ludwig-Maximilians-University}
  \city{Munich}
  \country{Germany}
}

\authornote{Both authors contributed equally to the paper}

\author{Jan N. van Rijn}
\affiliation{
  \institution{LIACS, Leiden University}
  \city{Leiden}
  \country{Netherlands}
 }
 
\author{Bernd Bischl}
\affiliation{
  \institution{Ludwig-Maximilians-University}
  \city{Munich}
  \country{Germany}
}

\author{Joaquin Vanschoren}
\affiliation{
  \institution{University of Eindhoven}
  \city{Eindhoven}
  \country{Netherlands}
}

\renewcommand{\shortauthors}{Gijsbers et al.}

\begin{abstract}
Hyperparameter optimization (HPO) in machine learning (ML) deals with the problem of empirically learning an optimal algorithm configuration from data, usually formulated as a black-box optimization problem. 
 In this work we propose a zero-shot method to meta-learn symbolic default hyperparameter configurations 
 that are expressed in terms of properties of the dataset. This enables a much faster, but still data-dependent configuration of the ML algorithm, compared to standard HPO approaches.
 In the past, symbolic and static default values have usually been obtained as hand-crafted heuristics.
 We propose an approach of learning such symbolic configurations as formulas of dataset properties, from a large set of prior evaluations of hyperparameters on multiple datasets and optimizing over a grammar of expressions by an evolutionary algorithm, similar to symbolic regression. We evaluate our method on surrogate empirical performance models as well as on real data across 6 ML algorithms on more than 100 datasets and demonstrate that our method can indeed find viable symbolic defaults.
\end{abstract}

\maketitle

\section{Introduction}
\label{sec:introduction}

The performance of most machine learning (ML) algorithms is greatly influenced by their hyperparameter settings. 
Various methods exist to automatically optimize hyperparameters, including random search~\cite{Bergstra2012}, Bayesian optimization~(\cite{Snoek2012,Hutter2011}), meta-learning~\cite{Brazdil2008} and bandit-based methods~\cite{Li2017}.
Depending on the algorithm, proper tuning of hyperparameters can yield considerable performance gains~\cite{Lavesson2006}. 
Despite the acknowledged importance of tuning hyperparameters, the additional run time, code complexity and experimental design questions cause many practitioners to often leave many hyperparameters at their default values, especially in real-world ML pipelines containing many hyperparameters. This is exacerbated by the fact that it is often unclear which hyperparameters should be tuned and which have negligible impact on performance (\cite{Probst2018}, \cite{probst2017}).
While not tuning hyperparameters can be detrimental, defaults provide a fallback for cases when results have to be obtained quickly, code complexity should be reduced by cutting out external tuning processes, or strong baseline results to compare to more complex tuning setups are needed.
Moreover, it seems less than ideal to optimize all hyperparameters from scratch with every new dataset. If the optimal values of a hyperparameter are functionally dependent on properties of the data, we could learn this functional relationship and express them as symbolic default configurations that work well across many datasets. That way, we can transfer information from previous optimization runs to obtain better defaults and better baselines for further tuning. 

This paper addresses a new meta-learning challenge: ``Can we \textit{learn} a vector of \textit{symbolic} configurations for multiple hyperparameters of state-of-the-art machine learning algorithms?'' Contrary to static defaults, symbolic defaults should be a function of the meta-features (dataset characteristics) of the dataset at hand, such as the number of features.  Ideally, these meta-features are easily computed, so that the symbolic defaults can be easily implemented into software frameworks with little to no computational overhead. 
Well-known examples for such symbolic defaults are already widely used: The random forest algorithm's default $mtry = \sqrt{p}$ for the number of features sampled in each split~\cite{rf_userguide}, the median distance between data points for the width\footnote{Or the inverse median for the inverse kernel width $\gamma$} of the Gaussian kernel of an SVM~\cite{sigest}, and many more. 
Unfortunately, it has not been studied, how such formulas can be obtained in a principled, empirical manner, especially when multiple hyperparameters interact, and have to be considered simultaneously.

\paragraph{Contributions} 
We propose an approach to learn such symbolic default configurations by optimizing over a grammar of potential expressions, in a manner similar to symbolic regression~\cite{koza94} using Evolutionary Algorithms.
We investigate how the performance of symbolic defaults compares to the performance of static defaults and simple optimization techniques such as random search on the possibly largest collection of metadata available.
We validate our approach across a variety of state-of-the-art ML algorithms and propose default candidates for use by practitioners.
In several scenarios, our procedure finds a symbolic configuration that outperforms static defaults, while on others, our procedure finds a static configuration containing only constants.
It should be noted that our approach is not limited to the algorithms studied in this paper. 
In fact, it is quite general, and could even be used for non-ML algorithms, as long as their performance is empirically measurable on instances in a similar manner.\\

The paper is structured as follows. After introducing relevant related work in Section \ref{sec:related}, we provide a motivating example in Section \ref{sec:motivating} to guide further intuition into the problem setting.
We then introduce and define the resulting optimization problem in Section \ref{sec:problem} before we continue with describing the proposed method in section \ref{sec:method}.
We study the efficacy of our approach in a broad set of experiments across multiple machine learning algorithms in Sections \ref{sec:expsetup} \& \ref{sec:results}.

\begin{figure}[t]
    \centering
    \begin{subfigure}[c]{0.45\textwidth}
    \includegraphics[width=\textwidth]{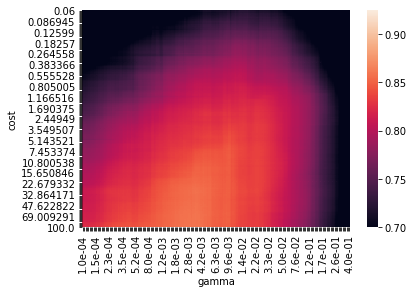}
    \subcaption{Linear cost and gamma}
    \end{subfigure}
    \begin{subfigure}[c]{0.45\textwidth}
    \includegraphics[width=\textwidth]{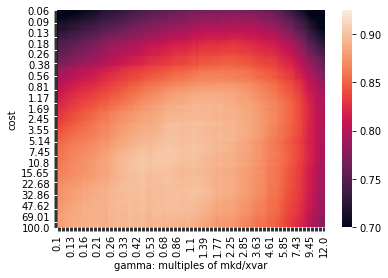}
    \subcaption{Linear cost, gamma multiple of mkd/xvar}
    \end{subfigure}
    \caption{Performance of an RBF-SVM averaged across 106 datasets for different values of cost and gamma, unscaled (a) and with gamma as multiples of meta-features (b)}
    \label{fig:svm_motivation}
\end{figure}

\section{Related Work}
\label{sec:related}

Symbolic defaults express a functional relationship between an algorithm hyperparameter value and dataset properties. Some example for such relationships are reported in literature, such as the previously mentioned formulas for the random forest~\cite{breiman01} or the SVM~\cite{sigest}. Some of these are also implemented in ML workbenches such as \texttt{sklearn} \cite{Pedregosa2011}, \texttt{weka} \cite{hall09_weka} or \texttt{mlr} \cite{mlr3}. It is often not clear and rarely reported how such relationships were discovered, nor does there seem to be a clear consensus between workbenches on which symbolic defaults to implement. Also, they are typically limited to a single hyperparameter, and do not take into account how multiple hyperparameters may interact.

Meta-learning approaches have been proposed to learn \textit{static} (sets of) defaults for machine learning algorithms  \cite{winkelmolen2020practical, mantovani2020rethinking, Pfisterer2018,Probst2018,Weerts2018,Wistuba2015learning} or neural network optimizers \cite{dl_multdefaults}, to analyze which hyperparameters are most important to optimize \cite{Rijn2018,Probst2018,Weerts2018}, or to build meta-models to select the kernel or kernel width in SVMs \cite{soares+04,valerio2014kernel,strang2018dont}.

An underlying assumption is that hyperparameter response surfaces across datasets behave similarly, and therefore settings that work well on some datasets also generalize to new datasets. Research conducted by warm-starting optimization procedures with runs from other datasets (c.f. \cite{lindauer-aaai18a}, \cite{feurerscalable}) suggest that this the case for many datasets.

Previous work \cite{Rijn2018MetaLF} on \textit{symbolic} defaults proposed a simplistic approach towards obtaining those, concretely by doing an exhaustive search over a space of simple formulas composed of an \textit{operator}, a \textit{numeric value} and a single \textit{meta-feature}. 
This significantly constricts the variety of formulas that can be obtained and might therefore not lead to widely applicable solutions. 

\todo{More background info on static search literature.}
\todo{Should we talk about EA literature?}

\section{Problem definition}
\label{sec:problem}

\subsection{Motivating Example} 
\label{sec:motivating}

The idea that reasonable hyperparameters settings can be described through a functional relationship based on dataset properties seems intuitive. 
We motivate this by considering the previously mentioned SVM example.
Figure~\ref{fig:svm_motivation} shows averaged response surfaces across $106$ tasks for hyperparameters $\gamma$ and $cost$ (zoomed in to a relevant area of good performance). While the scale for the cost parameter is kept fixed in Figures~\ref{fig:svm_motivation}(a) and \ref{fig:svm_motivation}(b), the x-axis displays the unchanged, direct scale for $\gamma$ in (a), and multiples of $\frac{mkd}{xvar}$ in (b) (symbols are described in Table \ref{tab:metafeatures}). This formula was found using the procedure that will be detailed in this paper. The maximum performance across the grid in (a) is $0.859$, while in (b) it is $0.904$. Values for $\frac{mkd}{xvar}$ range between $4.8\cdot10^{-5}$ and $0.55$. Empirically, we can observe several things. First, on average, a grid search over the $scaled$ domain in (b) yields better models. Secondly, the average solution quality and the area where good solutions can be obtained is larger, and the response surface is therefore likely also more amenable towards other types of optimization. And thirdly, we can conjecture that introducing formulas e.g. $\gamma = \frac{mkd}{xvar}$ for each hyperparameter can lead to better defaults. Indeed, finding good defaults in our proposed methodology essentially corresponds to optimization on an algorithm's response surface (averaged across several datasets). It should be noted that the manually defined heuristic used in \texttt{sklearn}, i.e. $\gamma = \frac{1}{p \cdot xvar}$, is strikingly similar.

 \subsection{Supervised Learning and Risk of a Configuration}

Consider a target variable $y$, a feature vector $\xv$, and an unknown joint distribution $\P$ on $(\xv,y)$, from which we have sampled a dataset $\D$ containing $N$ observations. An ML model $\fhx$ should approximate the functional relationship between $\xv$ and $y$.
An ML algorithm $\Al(\D)$ now turns the dataset (of size $N$) into a prediction model $\fhx$.
$\Al$ is controlled by a multi-dimensional hyperparameter configuration $\lv \in \Lambda$ of length $M$: $\lv = \{\lambda_1, \dots, \lambda_M\}$, where 
$\Lambda = \Lambda_1 \times \ldots \times \Lambda_M$ is a cross-product of (usually bounded) domains for all individual hyperparameters, so $\Lambda_j$ is usually a bounded real or integer interval, or a finite set of categorical values.  
In order to measure prediction performance pointwise between a true label and a prediction, we define a loss function $L(y, \hat{y})$.
We are interested in estimating the expected risk of the inducing algorithm w.r.t. $\lv$ on new data, also sampled from $\P$:
$$\RlP = E_\P(L(y, \Al(\D)(\xv))),$$
where the expectation above is taken over all 
datasets $\D$ of size $N$ from $\P$ and the test observation $(\xv, y)$.
Thus, $\RlP$ quantifies the expected predictive performance associated with a hyperparameter configuration $\lv$ for a given data distribution, learning algorithm and performance measure. 

\subsection{Learning an Optimal Configuration}

From a good \textit{default} configuration  $\lv$ we now expect that it performs well according to many of such risk mappings for many different data scenarios.
Given $K$ different datasets (or data distributions) $\P_1,...,\P_K$, we define $K$ hyperparameter risk mappings:
$$
\Rlk = E_{\P_k}(L(y, \Al(\D)(\xv))), \qquad k = 1,...,K.
$$

We now define the average risk of $\lv$ over $K$ data distributions: 
$$\Rl = \frac{1}{K}\sum_{k=1}^K \Rlk.$$

Minimizing the above w.r.t $\lv$ over $\Lambda$ defines an optimization problem for obtaining an optimal static configuration from $K$ scenarios, where we assume that, given a large enough $K$, a configuration will also work well on new data situations $\P$.

\subsection{Learning a Symbolic Configuration}

We now allow our configurations to be symbolic, i.e., contain formulas instead of static values. 
Hence, we assume that $\lv(.)$ is no longer a static vector from $\Lambda$, but a function that maps a dataset, or it's data characteristics, to an M-dimensional configuration.
$$\lv = (\lambda_1, \ldots, \lambda_M): \D \rightarrow \Lambda$$
For this reason, we define a context-free grammar of transformations, which define the space of potential expressions for all component functions $\lambda_j(.)$. 
This grammar consists of constant values, symbolic dataset meta-features and simple mathematical operators, detailed in  Table~\ref{tab:grammar}.

Given a meta-training set of $K$ data scenarios $\D_1, \ldots, \D_K$, 
we include the computation of the configuration by $\lv(\D)$ as a first step into the algorithm $\AlD$ and change our risk definition to: 
$$
\Rlk = E_{\P_k}(L(y, A_{\lambda(\D)}(\D)(\xv))), \qquad k = 1,...,K.
$$
and again average to obtain a global objective for $\lv(.)$: $$\Rl = \frac{1}{K}\sum_{k=1}^K \Rlk,$$
where the optimization now runs over the space of all potential M-dimensional formulas induced by our grammar.

\subsection{Metadata and Surrogates}

In principle, it is possible to estimate $\Rlk$ empirically using cross-validation during the optimization. However, this is obviously costly, as we want to obtain results for a large number of configurations across many multiple datasets. Therefore, we propose to employ \emph{surrogate models} that approximate $\Rlk$. 
We generate one surrogate for each dataset, ML algorithm and performance metric combination on a sufficiently large number of cross-validations experiments, with randomly planned design points for $\lv$.
Such meta-data evaluations are often used in literature (\cite{Wistuba2015, Wistuba2015learning, Rijn2018}), and can for example be obtained from \cite{Rijn2018, Kuehn2018} or \cite{Wistuba2015}.
This induces empirical surrogates, that map from static configurations to predicted performance values:
$$\hat{R}(\lv): \Lambda \rightarrow \reals$$
As our algorithm $\AlD$ is now removed, we simply change our objective to a simplified version, too:
$$\hat{R}(\lv(.)) = \frac{1}{K}\sum_{k=1}^K R_k(\lv(\D_k))$$
This defines a global, average risk objective for arbitrary formulaic $\lv(.)$ expressions that can be efficiently evaluated.

Considering the fact that performances on different datasets are usually not commensurable \cite{Demsar2006}, an appropriate scaling is required before training surrogate models to enable a comparison between datasets. This is done in literature by resorting to ranking~\cite{Bardenet13}, or scaling~\cite{yogatama14} to standard deviations from the mean. We mitigate the problem of lacking commensurability between datasets by scaling performance results to $[0;1]$ on a per-dataset basis as done in \cite{Pfisterer2018, dl_multdefaults}. After scaling, $1$ corresponds to the best performance observed in the meta-data and $0$ to the worst. A drawback to this is that some information regarding the absolute performance of the algorithm and the spread across different configurations is lost.

\paragraph{Dataset Characteristics}
In addition to the performance of random hyperparameter-configurations, OpenML contains a range of dataset characteristics, i.e, meta-features. A full list of available characteristics is described by~\cite{Rijn2016}. In order to obtain simple, concise and efficient formulas, we opted to include only simple dataset characteristics instead of working an extensive set as described by~\cite{Rijn2016}. 
Table \ref{tab:metafeatures} contains an overview over used meta-features and their corresponding ranges over our meta training set described in Section~\ref{sec:expsetup}.
Meta-features are computed for each dataset after imputation, one-hot encoding of categoricals and scaling of numeric features.
We include (among many others) the number of observations, the number of features and information regarding class balance. We denote the set of characteristics $\{c_1, c_2, ..., c_L\}$ with $C$. 
For this, we can also reuse evaluations shared on OpenML~\cite{Vanschoren2014}. 

\paragraph{Evaluation meta-data}
To learn symbolic defaults, we first gather meta-data that evaluates $R_k(\lv)$ on all $K$ datasets. For a given fixed algorithm with hyperparameter space $\Lambda$ and performance measure, e.g., logistic loss, a large number of experiments of randomly sampled $\lv$ is run on datasets $P_1, \dots, P_K$, estimating the generalization error of $\lv$ via 10-fold Cross-Validation.

\begin{table}[t]
    \centering
    \begin{tabular}{ll}
    \hline
    Symbol definition & Description \\
    \hline
    <configuration> ::= & \\
    $[$<F>|<I>$]$ * N & N: Number of hyperparameters \\
      & Type depends on hyperparameter \\
    \hline
    <I> ::= &  \\
      <unary> <F> & unary function \\
    | <binary> 2*<F> & binary function \\
    | <quaternary> 4*<F> & quaternary function \\
    | <i> & integer constant or symbol \\
    \hline
    <F> ::= & \\
    <I> & \\
    | <f> & float constant or symbol \\
    \hline
    <i> ::= & \\
    <mfi>  & Integer meta-feature, see Table~\ref{tab:metafeatures} \\
    | $c_i$ & $\lfloor x \rceil$; $x \sim$ loguniform($2^0, 2^{10}$) \\  
    \hline
    <f> ::= & \\
    <mff> & Continuous meta-feature, see Table~\ref{tab:metafeatures} \\
    | $c_f$ & $x \sim$ loguniform($2^{-10}, 2^0$)\\
    \hline
    <unary> ::= & \\
    exp & exp(x) \\
    | neg & -x \\
    \hline
    <binary> ::= & \\
    add & $x + y$ \\
    | sub & $x - y$ \\
    | mul & $x \cdot y$\\
    | truediv & $x / y$ \\
    | pow & $x^y$ \\
    | max & max($x, y$) \\
    | min & min($x, y$)\\
    \hline
    <quaternary> ::= & \\
    if\_greater & if a > b: c else d  \vspace{.5em} \\
    \hline
    \end{tabular}
    \caption{BNF Grammar for symbolic defaults search. <configuration> is the start symbol. }
    \label{tab:grammar}
\end{table}

\begin{table}[]
    \centering
    \begin{tabular}{llrrr}
    \toprule
    symbol &             explanation &    min &  median &       max \\
    \midrule
    <mfi>::=  &&&& \\
         n &         N. observations & 100 & 4147 & 130064 \\
        po &    N. features original &   4 &   36 &  10000 \\
         p &     N. features one-hot &   4 &   54 &  71673 \\
         m &              N. classes &   2 &    2 &    100 \\
    \hline \\
    <mff>::= &&&& \\
        rc &       N. categorical / p &  0.00 &    0.00 &      1.00 \\
      mcp &        Majority Class \% &   0.01 &    0.52 &      1.00 \\
      mkd &  Inv. Median Kernel Distance &   0.00 &    0.01 &      0.55 \\
      xvar &   Avg. feature variance &   0.00 &    1.00 &      1.00 \\
    \bottomrule
    \end{tabular}
    \caption{Available meta-features with corresponding symbols}
    \label{tab:metafeatures}
\end{table}

\section{Finding symbolic defaults}
\label{sec:method}
The problem we aim to solve requires optimization over a space of mathematical expressions. Several options to achieve this exist, e.g., by optimizing over a predefined fixed-length set of functions \cite{Rijn2018MetaLF}. 
One possible approach is to represent the space of functions as a grammar in Backus-Naur form and represent generated formulas as integer vectors where each entry represents which element of the right side of the grammar rule to follow \cite{Neil2001}. We opt for a tree representation of individuals, where nodes correspond to operations and leaves to terminal symbols or numeric constants, and optimize this via  genetic programming \cite{koza94}.
Our approach is inspired by symbolic regression \cite{koza_symbreg}, where the goal is to seek formulas that describe relationships in a dataset. In contrast, we aim to find a configuration (expressed via formulas), which minimizes $\hat{R}(\lv(.))$.

We differentiate between real-valued (<F>) and integer-valued (<I>) terminal symbols to account for the difference in algorithm hyperparameters. This is helpful, as real-valued and integer hyperparameters typically vary over different orders of magnitude. Simultaneously, some meta-features might be optimal when set to a constant, which is enabled through ephemeral constants.

A relevant trade-off in this context is the bias induced via a limited set of operations, operation depth and available meta-features. Searching, e.g., only over expressions of the form \textit{<binary>(<mff>, $c_f$)} introduces significant bias towards the form and expressiveness of resulting formulas.
Our approach using a grammar is agnostic towards the exact depth of resulting solutions, and bias is therefore only introduced via the choice of operators, meta-features and allowed depth.

Note that formulas can map outside of valid hyperparameter ranges. This complicates search on such spaces, as a large fraction of evaluated configurations might contain at least one infeasible setting. In order to reduce the likelihood of this happening, we define a set of mutation operators for the genetic algorithms that search locally around valid solutions. However if an infeasible setting is generated, the random forest surrogate effectively truncates it to the nearest observed value of that hyperparameter in the experiment meta-data, which is always valid.

Symbolic hyperparameters are interpretable and can lead to new knowledge i.e. about the interaction between dataset characteristics and performance.

\subsection{Grammar}

Table~\ref{tab:grammar} shows the primitives and non-symbolic terminals of the grammar, and Table~\ref{tab:metafeatures} shows the symbolic terminals whose values depend on the dataset. We define a set of \textit{unary}, \textit{binary} and \textit{quaternary} operators which can be used to construct flexible functions for the different algorithm hyperparameters.

The definition start symbol <configuration> indicates the type and number of hyperparameters available in a configuration and depends on the algorithm for which we want to find a symbolic default configuration.
In Table~\ref{tab:hpars} we indicate for each considered hyperparameter of a learner whether it is real-valued or integer-valued (the latter denoted with an asterisk ($^*$)).
For example, when searching for a symbolic default configuration for the decision tree algorithm, <configuration> is defined as <F><I><I><I>, because only the first hyperparameter is a float.
Expressions for integer hyperparameters are also rounded after their expression has been evaluated. 
Starting from <configuration>, placeholders <I> and <F> can now be iteratively replaced by either operators that have a return value of the same type or terminal symbols <i> and <f>. Terminal symbols can either be meta-features (c.f. Table \ref{tab:metafeatures}) or ephemeral constants.

\subsection{Algorithm}
\label{sec:alg}

We consider a genetic programming approach for optimizing the symbolic expressions.
We use a plus-strategy algorithm to evolve candidate solutions, where we set population size to 20 and generate 100 offspring in each generation via crossover and mutation. Evolution is run for $1000$ generations in our experiments.
We perform multi-objective optimization, jointly optimizing for performance of solutions (normalized logloss) while preferring formulas with smaller structural depth . Concretely, we employ NSGA-II selection \cite{deb2002fast} with binary tournament selection for parents and select offspring in an elitist fashion by usual non-dominated sorting and crowding distance as described in \cite{deb2002fast}.
For offspring created by crossover, the $M$ vector components of $\lv_j$ are chosen at random from both parents (uniform crossover on components), though we enforce at least one component from each parent is chosen.
This results in large, non-local changes to candidates.
Mutations, on the other hand, are designed to cause more local perturbations.
We limit their effect to one hyperparameter only, and include varying constant values, pruning or expanding the expression or replacing a node.
Each offspring is created through either crossover or mutation, never a combination.
More details w.r.t. employed mutations can be found in Appendix~\ref{sec:appendix_impl}.
Initial expressions are generated with a maximum depth of three.
We do not limit the depth of expressions during evolution explicitly, though multi-objective optimization makes finding deep formulas less likely. 

\section{Experimental Setup}

\label{sec:expsetup}
We aim to answer the following research questions:\\
\textbf{RQ1}: How good is the performance of symbolic defaults in a practical sense? In order to asses this, we compare symbolic defaults with the following baselines: $a)$ existing defaults in current software packages $b)$ static defaults found by the search procedure described in Section \ref{sec:method}, disallowing meta-features as terminal symbols and $c)$ a standard hyperparameter random search (on the surrogates) with different budgets. Note that existing defaults already include symbolic hyperparameters in several implementations. In contrast to defaults obtained from our method, existing implementation defaults are often not empirically evaluated, and it is unclear how they were obtained. 
This question is the core of our work, as discrepancy to evaluations on real data only arise from inaccurate surrogate models, which can be improved by tuning or obtaining more data.\\
\textbf{RQ2}: How good are the symbolic defaults we find, when evaluated on real data? We evaluate symbolic defaults found by our method with experiments on real data and compare them to existing implementation defaults and a simple meta-model baseline.

\subsection{General setup}
\begin{table}
    \centering
    \begin{tabular}{lp{30mm}p{30mm}}
    \hline
    algorithm & fixed & optimized \\
    \hline
    elastic net & - & $\alpha$, $\lambda$ \\
    \hline
    decision tree & - & $cp$, $max.depth^*$ , $minbucket^*$, $minsplit^*$\\
    \hline
    random forest & splitrule:gini,
    num.trees:500,\hspace{1cm} replace:True & $mtry^*$, $sample.fraction$, $min.node.size^*$ \\
    \hline
    svm & kernel:radial & $C$, $\gamma$\\
    \hline
    approx. knn & distance:l2 & $k^*$, $M^*$, $ef^*$, $efc^*$ \\
    \hline
    xgboost & booster:gbtree& 
        $\eta$, $\lambda$, $\gamma$, $\alpha$, $subsample$,
        $max\_depth^*$, $min\_child\_weight$, $colsample\_bytree$, $colsample\_bylevel$\\
    \hline
    \end{tabular}                       
    \caption{Fixed and optimizable hyperparameters for different algorithms. Hyperparameters with an asterisk (*) are integers.}
    \label{tab:hpars}
\end{table}

We investigate symbolic defaults for $6$ ML algorithms using the possibly largest available set of meta-data, containing evaluations of between $106$ and $119$ datasets included either in the OpenML-CC18~\cite{Bischl2017} benchmark suite or the AutoML benchmark \cite{Gijsbers2019AnOS}. Datasets have between $100$ and $130000$ observations, between $3$ and $10000$ features and $2-100$ classes. The number of datasets varies across algorithms as we restrict ourselves to datasets where the experimental data contains evaluations of at least $100$ unique hyperparameter configurations available as well as surrogate models that achieve sufficient quality (Spearman's $\rho > 0.8$). We investigate implementations of a diverse assortment of state-of-the-art ML algorithms, namely support vector machines \cite{cortes1995support}, elastic net \cite{zou2005regularization}, approximate knn \cite{approxknn}, decision trees \cite{cart}, random forests \cite{breiman01} and extreme gradient boosting (xgboost, \cite{xgboost}). The full set of used meta-features can be obtained from Table \ref{tab:metafeatures}. The choice of datasets and ML algorithms evaluated in our paper was based on the availability of high-quality metadata.
A large number of random evaluations across the full configuration space of each algorithm for each dataset was obtained and used in order to fit a random forest surrogate model for each dataset / algorithm combination\footnote{\url{https://www.openml.org/d/4245[4-9]}}.
We optimize the average logistic loss across 10 cross-validation folds (normalized to [0,1]), as it is robust to class-imbalancies, but our methodology trivially extends to other performance measures. The hyperparameters optimized for each algorithm are shown in Table \ref{tab:hpars}. For the random forest, we set the number of trees to $500$ as recommended in literature \cite{probst_tune_rf}. We perform $10$ replications of each experiment for all stochastic algorithms and present aggregated results.

\subsection{Experiments for RQ1 \& RQ2}

The evaluation strategy for both experiments is based on a leave-one-data-set-out strategy, where the (symbolic) defaults are learned on all but one dataset, and evaluated using the held-out dataset.\\
In \textit{Experiment 1} for \textbf{RQ1}, we evaluate symbolic defaults found using our approach against baselines mentioned in \textbf{RQ1}. Our hold-out-evaluation is performed on a held-out surrogate.\\
In \textit{Experiment 2} for \textbf{RQ2}, now learn our symbolic defaults in the same manner as for \textit{Experiment 2} on $K-1$ surrogates, but now evaluate their performance via a true cross-validation on the held out dataset instead of simply querying the held out surrogate.\\

Our main experiment -- Experiment 1 evaluates symbolic defaults on surrogates for a held-out task, which let's us measure whether our symbolic defaults can extrapolate to future datasets.
If results from surrogate evaluation correspond to real data, we can also conclude that our surrogates approximate the relationship between hyperparameters and performance well enough to transfer to real-world evaluations. We conjecture, that, given the investigated search space, for some algorithms/hyperparameters symbolic defaults do not add additional benefits and constant defaults suffice. In those cases, we expect that our approach performs roughly as well as an approach that only takes into account constant values, because our approach can similarly yield purely constant solutions.

We employ a modified procedure, \textit{optimistic random search} that simulates random search on each dataset which is described below. We consider random search with budgets of up to 32 iterations to be strong baselines. Other baselines like Bayesian optimization are left out of scope, as we only evaluate a single symbolic default, which is not optimized for the particular dataset.
In contrast to requiring complicated evaluation procedures such as nested cross-validation, our symbolic defaults can simply be implemented as software defaults. We further consider full AutoML systems such as auto-sklearn \cite{Feurer2015} to be out-of-scope for comparison, as those evaluate across pre-processing steps and various ML algorithms, while our work focuses on finding defaults for a single algorithm without any pre-processing.

\paragraph{Optimistic random search}
\label{sec:ors}
As we deal with random search in the order of tens of evaluations, obtaining reliable results would require multiple replications of random search across multiple datasets and algorithms. We therefore adapt a cheaper, optimistic random search procedure, which samples $budget$ rows from the available metadata for a given dataset, computes the best performance obtained and returns it as the random search result. Note that this assumes that nested cross-validation performance generalizes perfectly to the outer cross-validation performance, which is why it is considered an optimistic procedure. It is therefore expected to obtain higher scores than a realistic random search would obtain. Nonetheless, as we will show, will the single symbolic default often outperform the optimistic random search procedure with 8-16 iterations. The optimistic random search procedure is repeated several times in order to obtain reliable estimates.\\

\paragraph{1-Nearest Neighbour}
\label{sec:1nn}
In Experiment 2, where we conduct evaluations with 10-fold cross-validation on the data, we also compare to a simple meta-model for generating a candidate solution given the meta-dataset. We use the k-Nearest Neighbour approach used by auto-sklearn~\cite{Feurer2015}, which looks up the best known hyperparameter configuration for the \textit{nearest} dataset. To find the nearest datasets each meta-feature is first normalized using min-max scaling, then distances to each dataset are computed using $L_1$-norm. While auto-sklearn finds hyperparameter configuration candidates for each of the 25 nearest neighbours, we only use the best hyperparameter configuration from the first nearest neighbour.\\

\noindent The Python code for our experiments is available online 
\footnote{\url{https://github.com/PGijsbers/symbolicdefaults}}
and makes use of the DEAP module~\cite{DEAP_JMLR2012} for genetic programming.

\section{Results}
\label{sec:results}
In this section, we will first analyse the quality of our surrogates models in order to ensure their reliability. Next, we evaluate the performance of the found symbolic defaults on both surrogates and real data.

\subsection{Surrogates and Surrogate Quality}
As described earlier, we use surrogate models to predict the performance of hyperparameter configurations to avoid expensive evaluations.
It is important that the surrogate models perform well enough to substitute for real experiments.
For this optimization task, the most important quality of the surrogate models is the preservation of relative order of hyperparameter configuration performance.
Error on predicted performance is only relevant if it causes the optimization method to incorrectly treat a worse configuration as a better one (or vice versa).
The performance difference itself is irrelevant.

\begin{figure}
    \centering
    \begin{subfigure}[c]{0.23\textwidth}   
    \includegraphics[width=\textwidth]{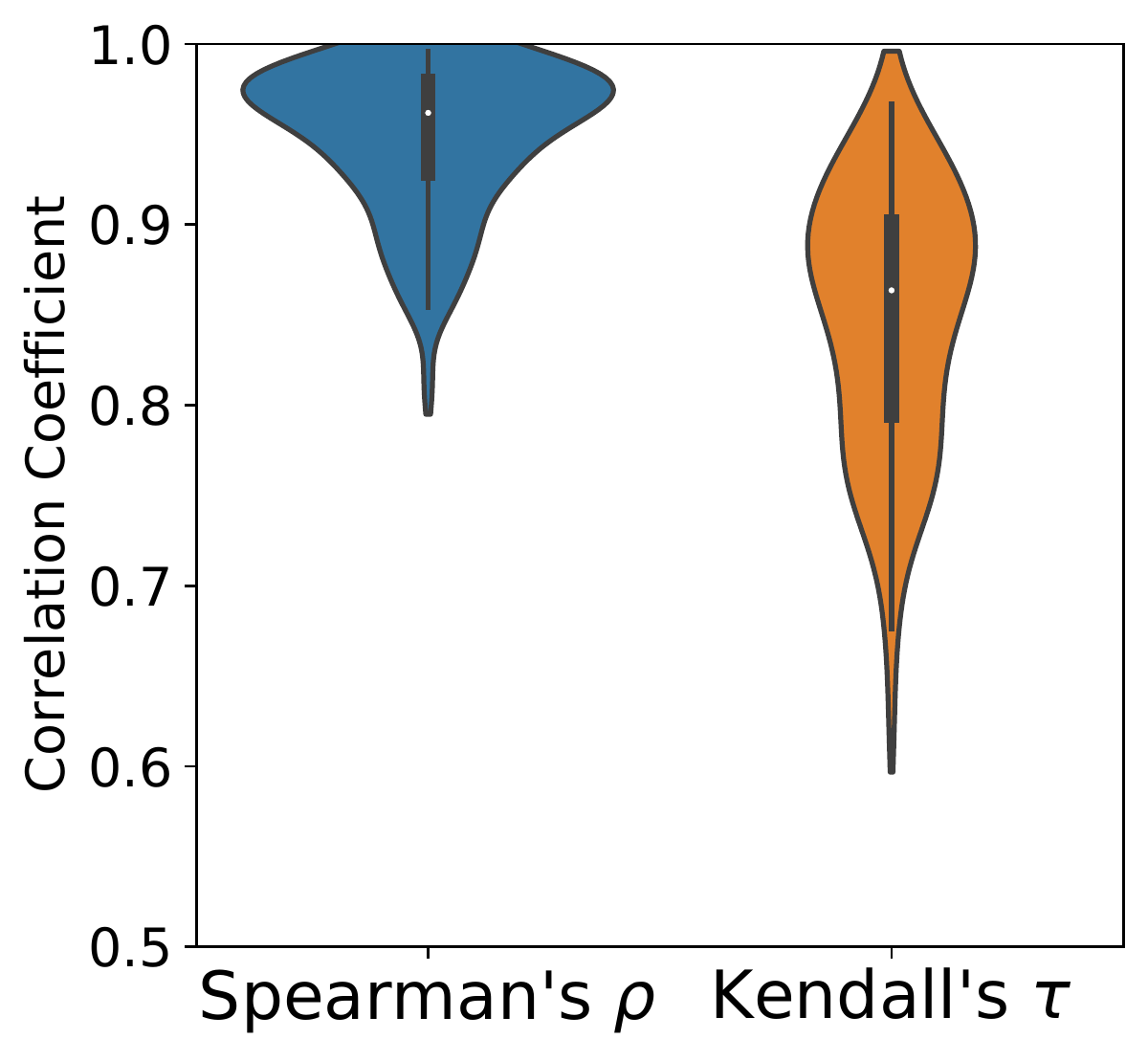}
    \end{subfigure}
    \begin{subfigure}[c]{0.23\textwidth}
    \includegraphics[width=\textwidth]{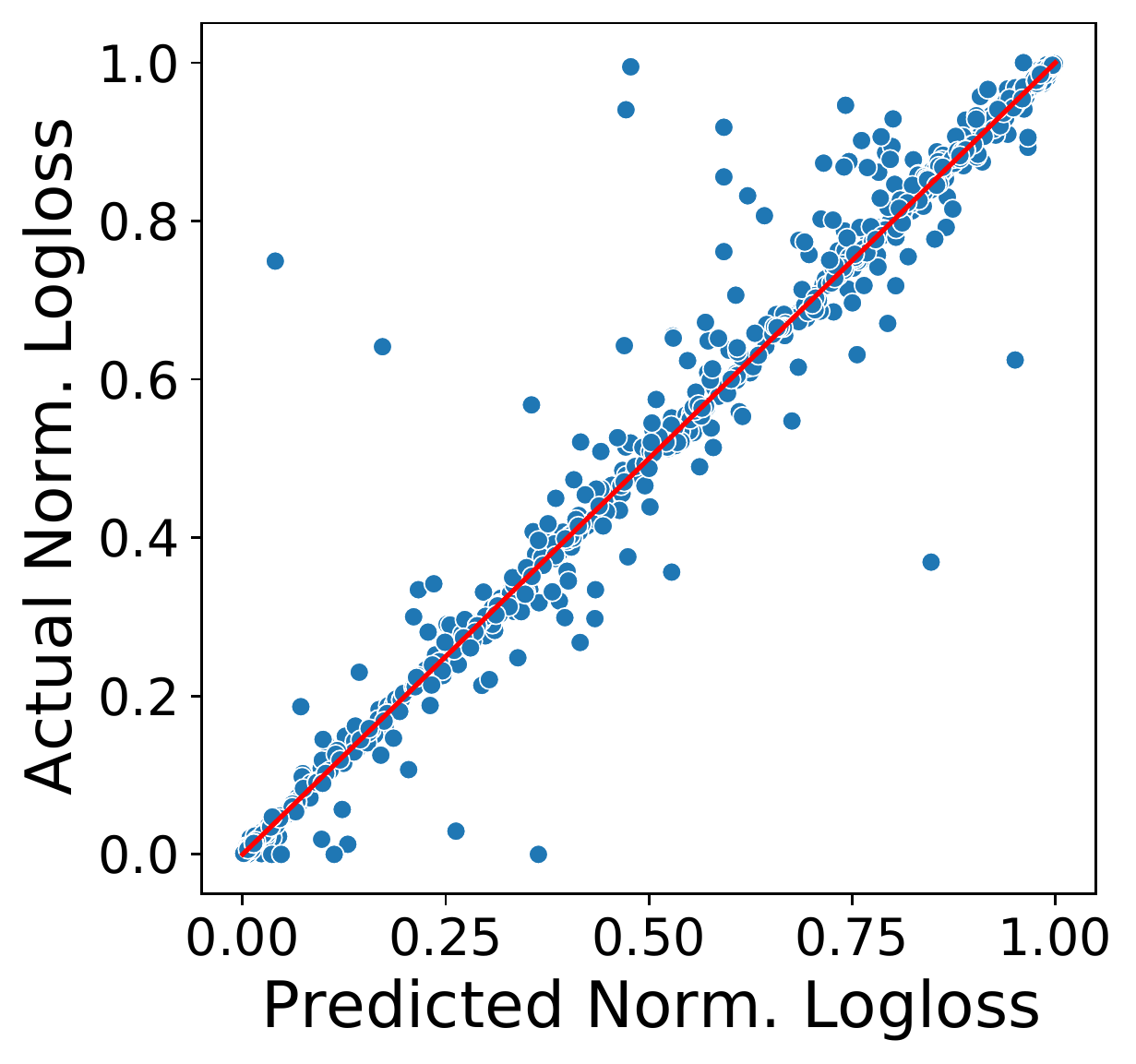}
    \end{subfigure}
    \caption{Comparison of performance predicted by the SVM surrogate against the real performance across tasks. On the left by their rank correlation coefficients, on the right in normalized performance.}
    \label{fig:surrogate_quality}
\end{figure}

For that reason, we evaluate our surrogate models first on rank correlation coefficients Spearman's $\rho$ and Kendall's $\tau$.
For each task, we perform 10-fold cross validation on our meta data, and calculate the rank correlation between the predicted ranking and the true one.
On the left in Figure~\ref{fig:surrogate_quality} we show the distribution of Spearman's $\rho$ and Kendall's $\tau$ across 106 tasks for the SVM surrogate.
Due to the high number of observations all rank correlations have a p-value of near zero.
We observe high rank correlation for both measurements on most tasks.
That $\tau$ values are lower than $\rho$ values indicates that the surrogate model is more prone to making small mistakes than big ones.
This is a positive when searching for good performing configurations, but may prove detrimental when optimizing amongst good configurations.

While it does not directly impact search, we also look at the difference between the predicted and true performances.
For each task 10 configurations were sampled as a test set, and a surrogate model was trained on the remainder of the meta data for that task.
On the right in Figure~\ref{fig:surrogate_quality} the predicted normalized performance is shown against the real normalized performance.
Predictions closer to the diagonal line are more accurate, points under and over the diagonal indicate the surrogate model was optimistic and pessimistic respectively.
Plots for the other models can be found in Appendix~\ref{app:exps_surr}.

\subsection{Experiment 1 - Benchmark on surrogates} 
\label{sec:exps_rq2}
In order to answer \textbf{RQ1}, we compare the performance of symbolic defaults, constant defaults and existing implementation defaults on surrogates. Implementation defaults are default values currently used for the corresponding algorithm implementations and can be obtained from Table \ref{tab:baseline_defaults} in the appendix. Note that random search in this context does refer to per-task optimistic random-search as described in \ref{sec:expsetup}. In the following, we analyze results for the SVM and report normalized out-of-bag logistic loss on a surrogate if not stated otherwise. We conduct an analysis of all other algorithms mentioned in \ref{tab:hpars} in Appendix~\ref{app:exps_surr} a-e).

\begin{figure}
    \centering
    \includegraphics[width=0.48\textwidth]{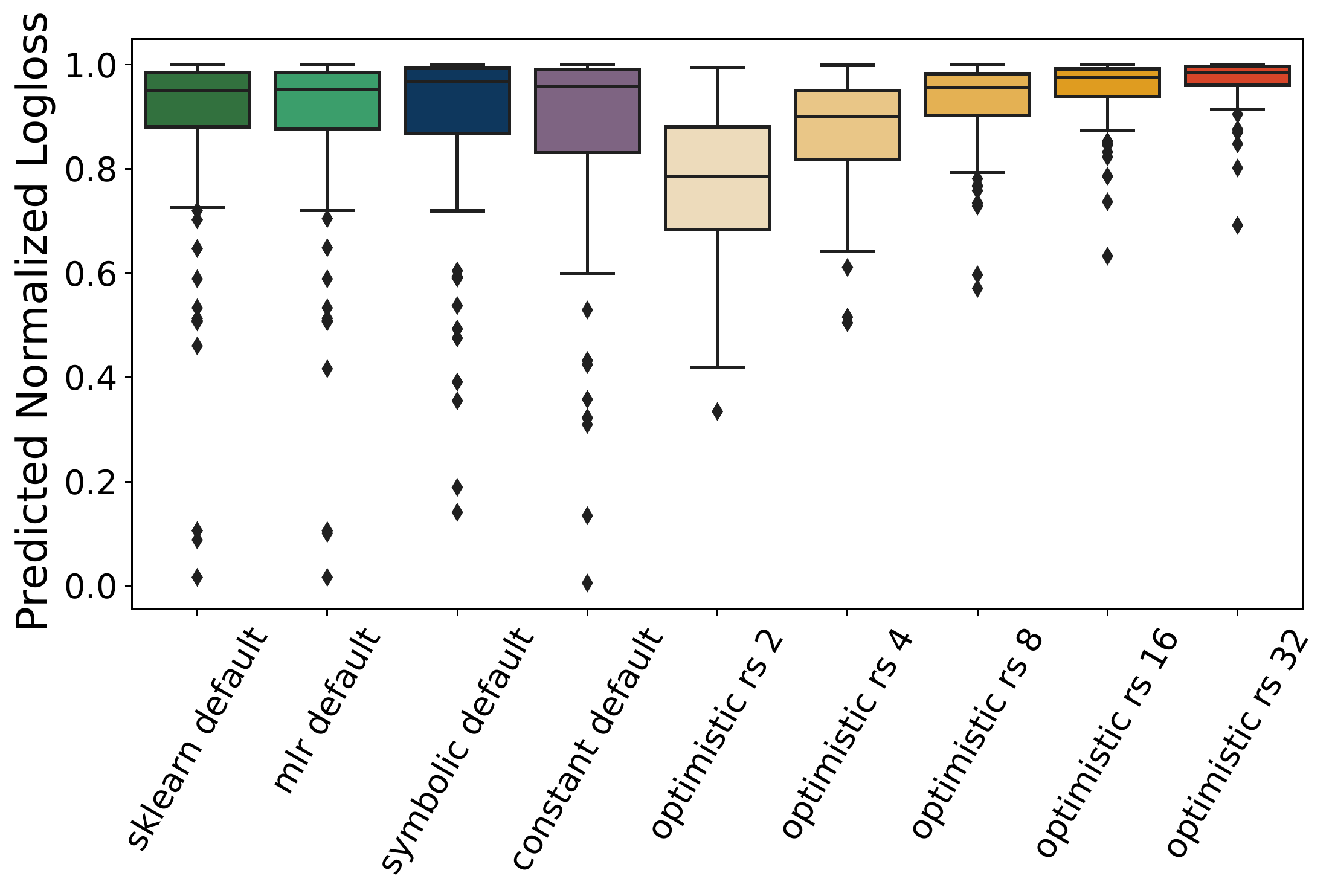}
    \caption{Symbolic, static and implementation defaults for SVM, comparing normalized logloss predicted by surrogates.}
    \label{fig:boxplot_surrogates}
\end{figure}

A comparison to baselines $a-c)$ for the SVM can be obtained from Figure \ref{fig:boxplot_surrogates}. We compare symbolic defaults (blue), existing implementation defaults (green), constant defaults (purple) and several iterations of random search (orange). 
Symbolic defaults slightly outperform existing implementation defaults (mlr default and sklearn default) and compare favorably to random search with up to $8$ evaluations. 

\begin{figure}
    \centering
    \includegraphics[width=0.48\textwidth]{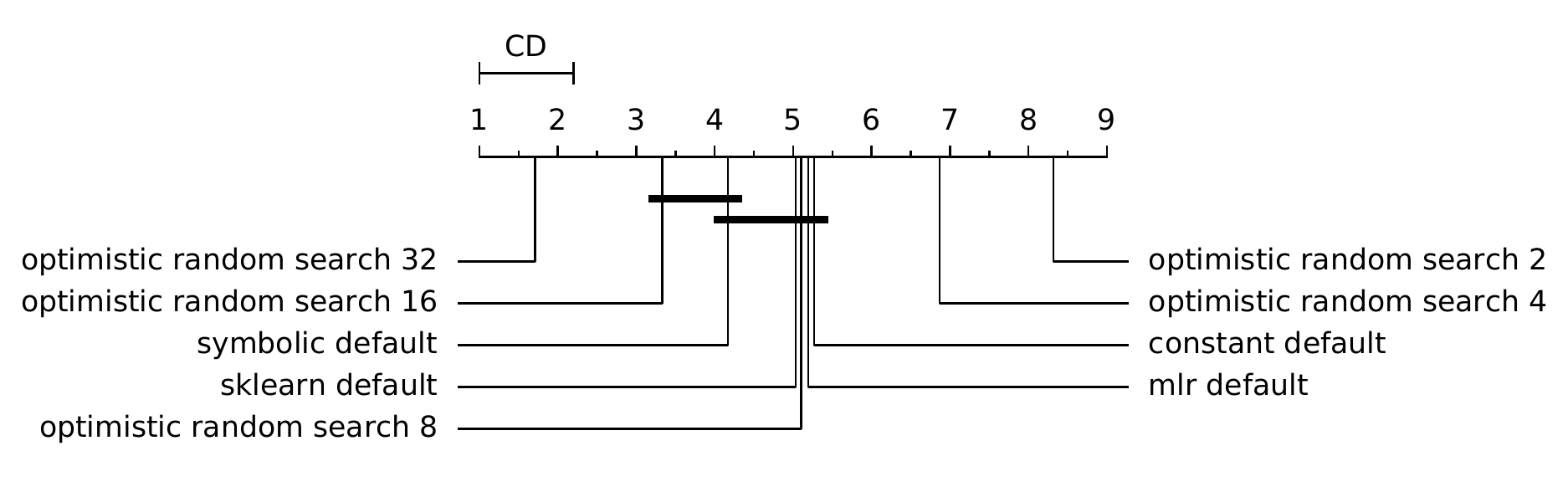}
    \caption{Critical Differences Diagram for symbolic, static and implementation defaults on surrogates}
    \label{fig:cd_surrogates}
\end{figure}

For significance tests we use a non-parametric Friedman test for differences in samples at $\alpha=0.05$ using  and a post-hoc Nemenyi test. The corresponding critical differences diagram \cite{Demsar2006} is displayed in Figure \ref{fig:cd_surrogates}. Methods are sorted on the x-axis by their average rank across tasks (lower is better). For methods connected by a bold bar, significant differences could not be obtained. Symbolic defaults do not perform significantly worse than random search with a budget of $16$ evaluations, however they also do not significantly outperform the hand-crafted implementation defaults or an optimized constant default.

\begin{figure}
    \centering
    \begin{subfigure}[c]{0.23\textwidth}
    \includegraphics[width=\textwidth]{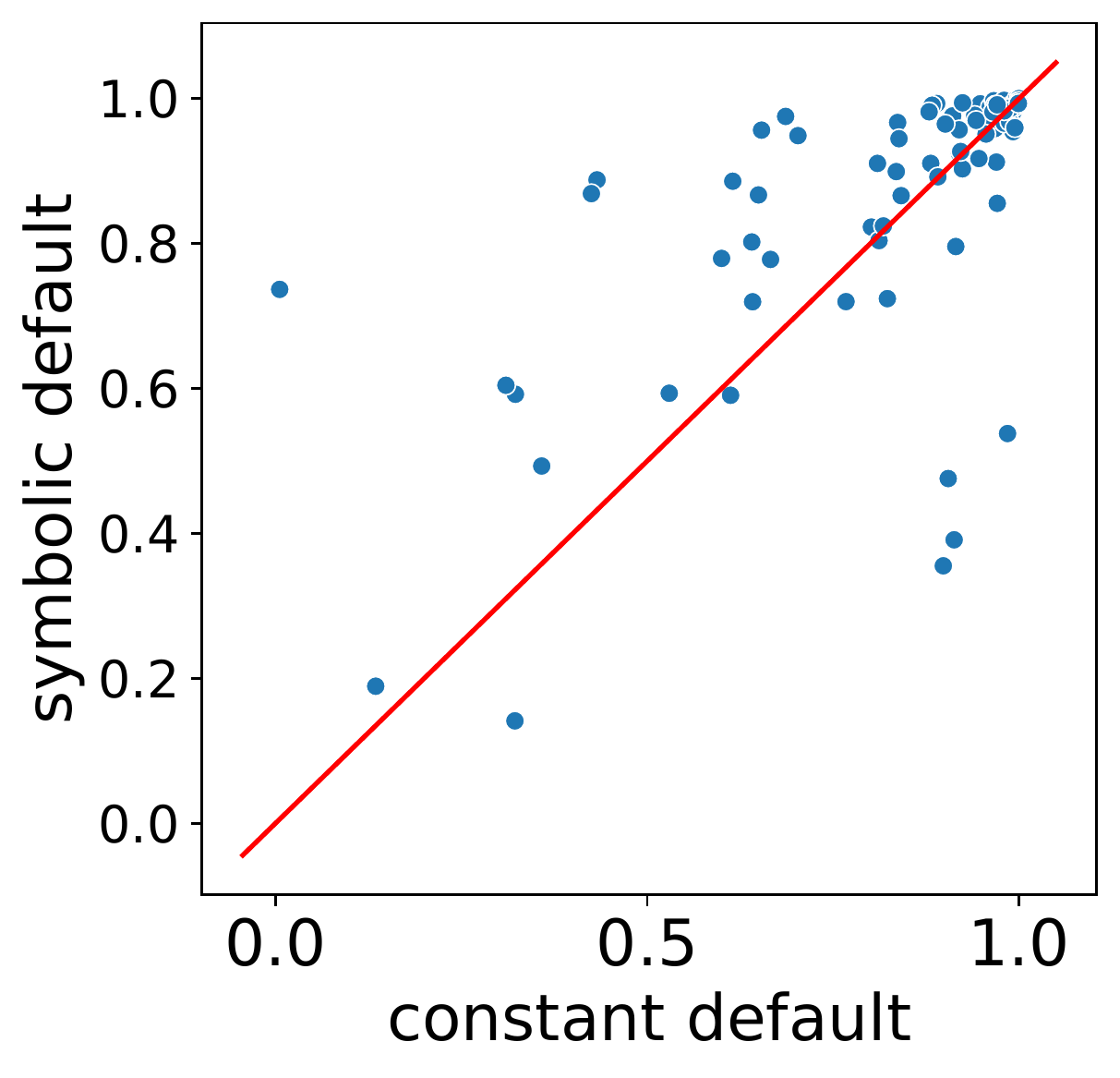}
    \end{subfigure}
    \begin{subfigure}[c]{0.23\textwidth}
    \includegraphics[width=\textwidth]{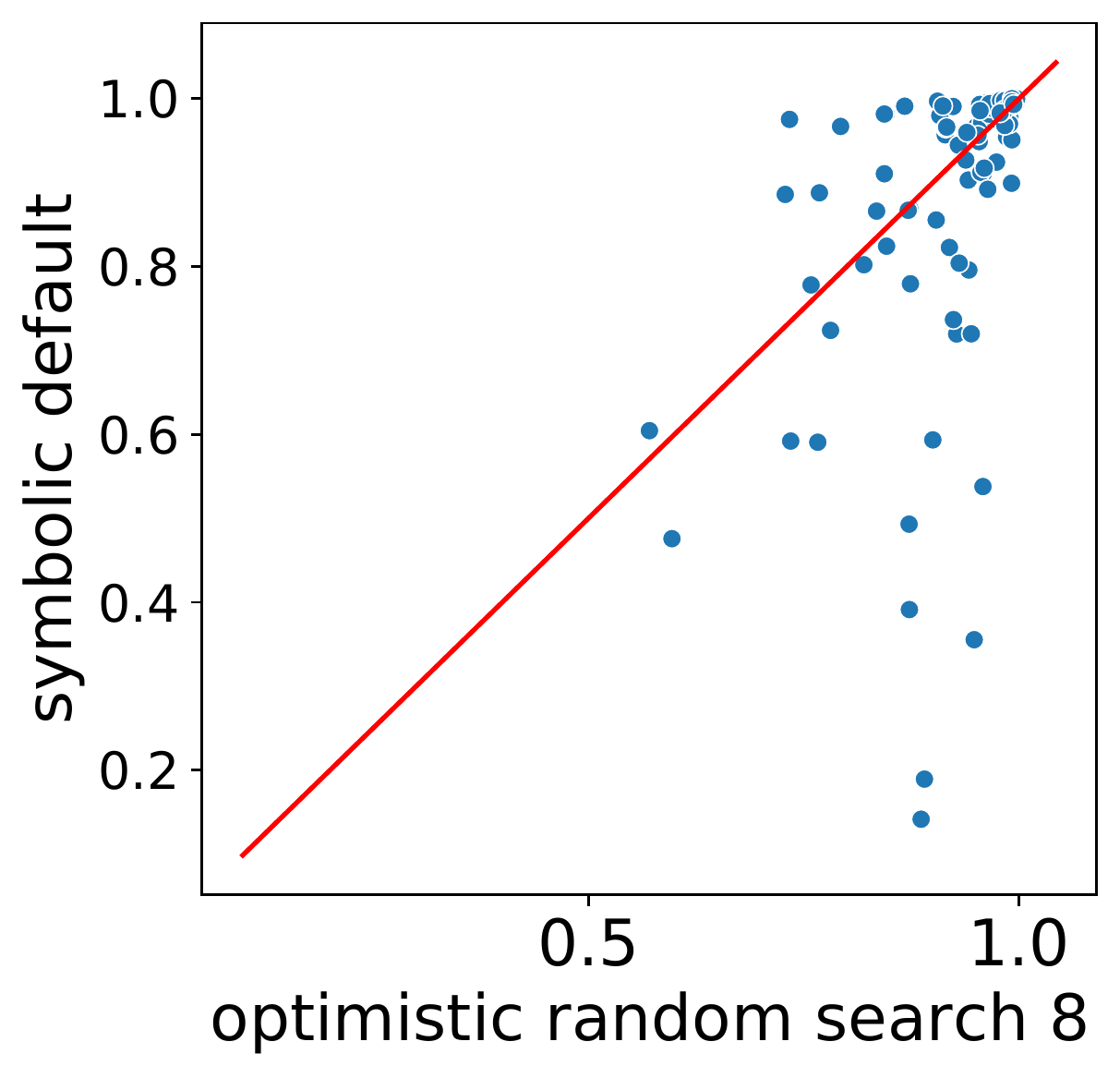}
    \end{subfigure}
    \caption{Performance comparison of symbolic defaults to constant defaults (left) and budget 8 random search (right). Points above the red line indicate symbolic defaults are better.}
    \label{fig:scatter_svm_surrogates}
\end{figure}

Figure \ref{fig:scatter_svm_surrogates} shows comparisons to baselines, again using normalized logistic loss. The y-axis in both cases corresponds to symbolic defaults, while the x-axis corresponds to constant defaults (left) and random search with a budget of $8$ (right).
We conduct a similar analysis on all other algorithms in the appendix.

We summarize the results across all experiments in Table \ref{tab:surrogate_results}, which shows the mean normalized logistic loss and standard deviation across all tasks for each algorithm. The symbolic and constant column denote the performance of defaults found with our approach including and excluding symbolic terminals respectively. The package column shows the best result obtained from either the scikit-learn or mlr default, and the last column denotes the best found performance sampling 8 random real world scores on the task for the algorithm.

We find that the symbolic default mean rank is never significantly lower than that of other approaches, but in some cases it is significantly higher (in bold). While the mean performance for symbolic solutions is lower for glmnet, random forest and rpart, we observe that the average rank is only higher for glmnet (see Section~\ref{sec:exps_surr}).

The only implementation default which does not score a significantly lower mean rank than the defaults found by search with symbolic terminals is the default for SVM, which has carefully hand-crafted defaults. This further motivates the use of experiment data for tuning default hyperparameter configurations. In three out of six cases the tuned defaults even outperform eight iterations of the optimistic random search baseline, in the other cases they have a significantly higher mean rank than 4 iterations of random search.

\begin{table}[h]
    \centering
    \resizebox{0.49\textwidth}{!}{%
    \begin{tabular}{lrrrr}
    \toprule
    algorithm &  symbolic &  constant &  package  &  opt. RS 8 \\
    \midrule
    glmnet  &             \textbf{0.917(0.168)} &             \textbf{0.928(0.158)} &                   0.857(0.154) &                       0.906(0.080) \\
    knn     &             \textbf{0.954(0.148)} &             0.947(0.156) &                   0.879(0.137) &                       \textbf{0.995(0.009)} \\
    rf      &             \textbf{0.946(0.087)} &             0.951(0.074) &                   0.933(0.085) &                       0.945(0.078) \\
    rpart   &             \textbf{0.922(0.112)} &             \textbf{0.925(0.093)} &                   0.792(0.141) &                       \textbf{0.932(0.082)} \\
    svm     &             \textbf{0.889(0.178)} &             \textbf{0.860(0.207)} &                   \textbf{0.882(0.190)} &                       \textbf{0.925(0.084)} \\
    xgboost &             \textbf{0.995(0.011)} &             \textbf{0.995(0.011)} &                   0.925(0.125) &                       0.978(0.043) \\
    \bottomrule
    \end{tabular}}
    \caption{Mean normalized logloss and standard deviation (in parentheses) across all tasks and comparison to baselines $a-c)$ across all learners. \textit{Package} denotes the best of scikit-learn and mlr defaults. Bold indicates the average rank was not significantly worse than the best of the four defaults.}
    \label{tab:surrogate_results}
\end{table}

\subsection{Experiment 2 - Benchmark on real data} 
\label{sec:exps_rq3}

We run the defaults learned on $K-1$ surrogates for each hold-out dataset with a true cross-validation and compare its performance to existing implementation defaults. We again analyze results for SVM and provide results on other algorithms in the appendix.
Note, that instead of normalized log-loss (where 1 is the optimum), we report standard log-loss in this following section, which means lower is better. Figure ~\ref{fig:svm_realdata} shows box plot and scatter plot comparisons between the better implementation default (sklearn) and symbolic defaults obtained from our method. The symbolic defaults found by our method performs slightly better to the two existing baselines in most cases, but outperforms the sklearn default on some datasets while never performing drastically worse. This small difference might not be all-too-surprising, as the existing sklearn defaults are already highly optimized symbolic defaults in their second iteration \cite{sklearndocs}.

\begin{figure}
    \centering
\begin{subfigure}[c]{0.24\textwidth}
    \centering
    \includegraphics[width=\textwidth]{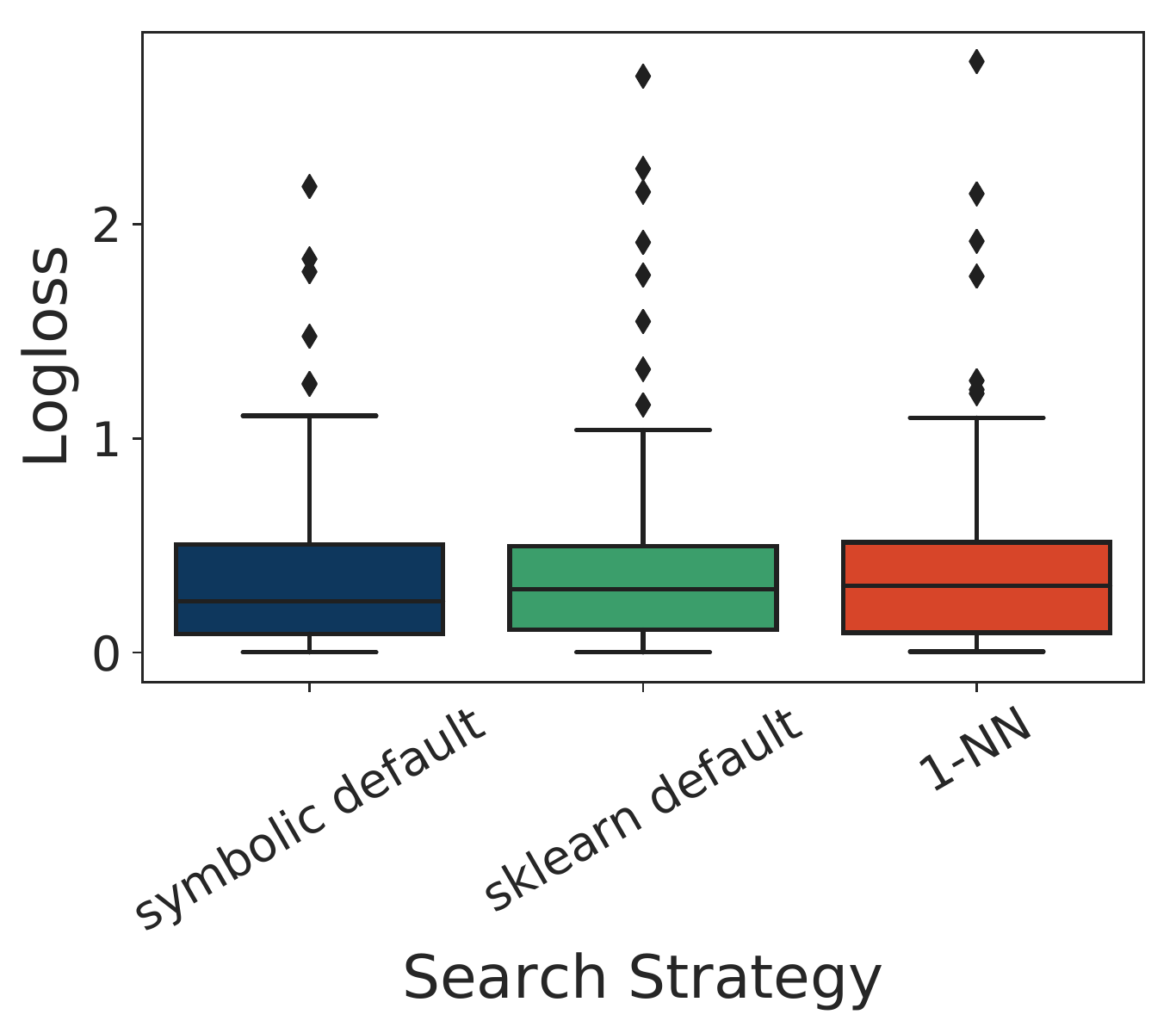}
\end{subfigure}
\begin{subfigure}[c]{0.23\textwidth}
    \centering
    \includegraphics[width=\textwidth]{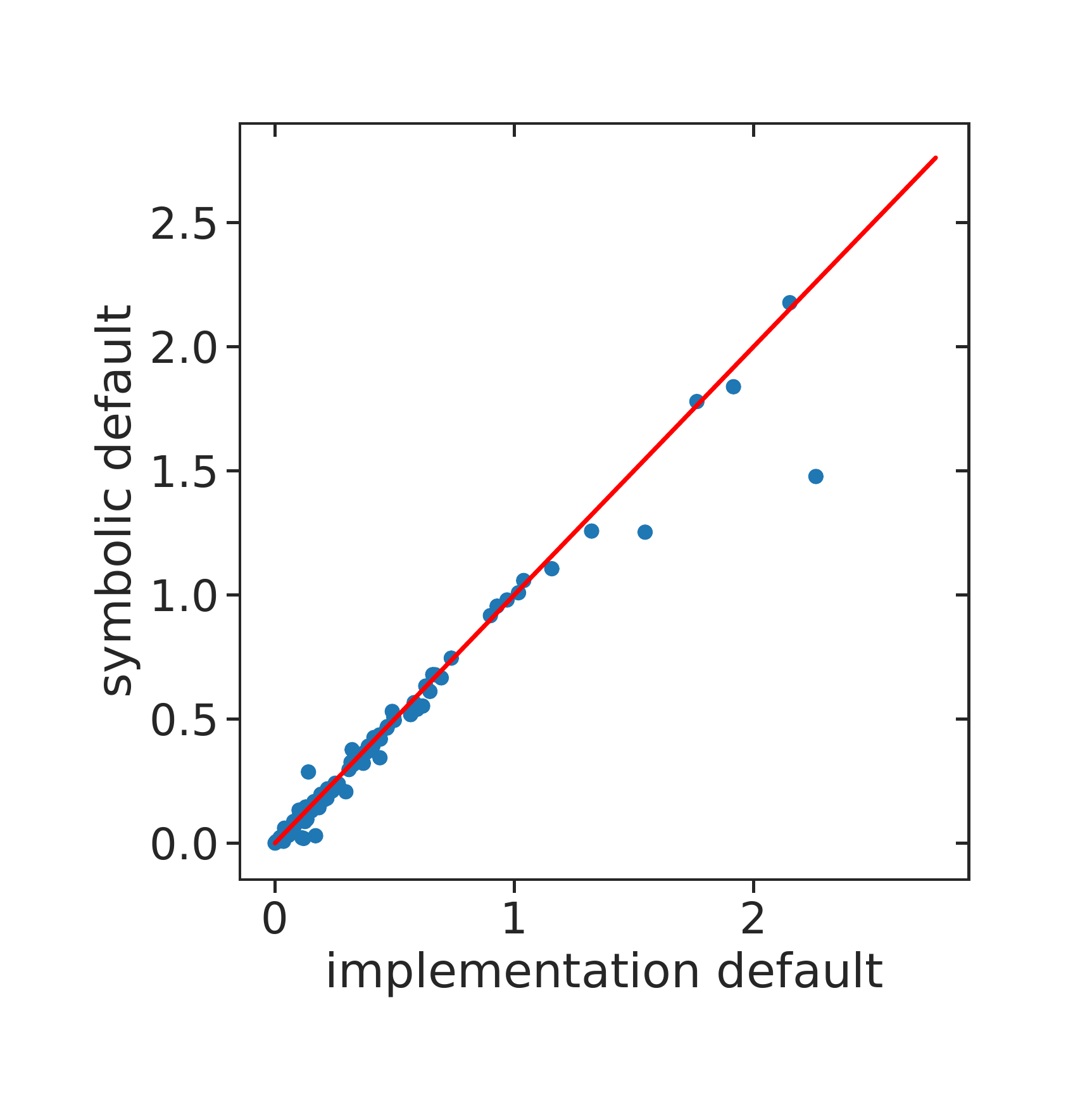}
\end{subfigure}
    \caption{Comparison of symbolic and implementation default using log-loss across all datasets performed on real data. Box plots (right) and scatter plot (left)}
    \label{fig:svm_realdata}
\end{figure}

\section{Conclusion and Future Work}

In this paper we consider the problem of finding data-dependent hyperparameter configurations that work well across datasets.
We define a grammar that allows for complex expressions that can use data-dependent meta-features as well as constant values.
Surrogate models are trained on a large meta dataset to efficiently optimize over symbolic expressions.

We find that the data-driven approach to finding default configurations leads to defaults as good as hand-crafted ones. The found defaults are generally better than the defaults set by algorithm implementations. Depending on the algorithm, the found defaults can be as good as performing 4 to 16 iterations of random search.
In some cases, defaults benefit from being defined as a symbolic expression, i.e. in terms of data-dependent meta-features.

In future work we first of all plan to extend the search space and extend single symbolic configurations to sets of symbolic configurations. 
We aim to extend the search space in two ways: dataset characteristics have to reflect properties that are relevant to the algorithm hyperparameters, yet it is not immediately clear what those relevant properties are. It is straightforward to extend the number of meta-features, as many more have already been described in literature (c.f. \cite{Rijn2016}). This might not only serve to find even better symbolic defaults, but also further reduces bias introduced by the small number of dataset characteristics considered in our work.
By extending the grammar described in Table \ref{tab:grammar} to include categorical terminals and operators more suitable for categorical hyperparameters (e.g. \textit{if-else}), the described procedure can extend to categorical and hierarchical hyperparameters. This could also be extended to \textit{defaults across algorithms}, i.e., given a set of dataset characteristics, which algorithm and which hyperparameter should be used as a default. 

Furthermore, we aim to extend the notion of defaults to sets of defaults as done in \cite{Pfisterer2018} and \cite{dl_multdefaults}. Both propose sets of defaults, which can serve as good starting points for evaluation and further hyperparameter tuning. Using evolutionary algorithms, the greedy forward search used to find optimal sets of solutions can be forgone in favour of multi-objective strategies based on \textit{cooperative co-evolution}, where archives of solutions for each task are co-evolved(~\cite{mo_coevolution}, \cite{innovation_engines} \cite{CMOEA}). This can lead to smaller, jointly optimal sets of default algorithms.

Another relevant aspect, which we do not study in this work is the \textit{runtime} associated with a given default, as we typically want default values to be fast as well as good, and therefore this trade-off might be considered in optimizing symbolic defaults.
In this work, we address this by restricting specific hyperparameters to specific values, in particular the xgboost \texttt{nrounds} parameter to $500$. In future research, we aim to take this into consideration for all methods.
Having access to better, symbolic defaults, makes machine learning more accessible and robust to researchers from all domains. 

\begin{acks}
This material is based upon work supported by the Data Driven Discovery of Models (D3M) program run by DARPA and the Air Force Research Laboratory, and by the German Federal Ministry of Education and Research (BMBF) under Grant No. 01IS18036A. The authors of this work take full responsibilities for its content.
\end{acks}

\bibliographystyle{acmref}
\bibliography{refs}

\clearpage
\newpage

\appendix

\section{Implementation defaults}
\label{sec:appendix_impl}

\begin{table}[b]
    \centering
    \begin{tabular}{p{15mm}l}
    \toprule
    Algorithm & Default\\
    \midrule
    Elastic Net & 
    \begin{tabular}{p{15mm}p{35mm}}
        glmnet: &  $\alpha:1$, $\lambda:0.01$ \\
    \end{tabular}\\
    \hline
    Decision Tree &
    \begin{tabular}{p{15mm}p{45mm}}
        rpart: &  $cp:0.01$, $max.depth:30$, $minbucket:1$, $minsplit:20$ \\
    \end{tabular}\\
    \hline
    Random Forest & 
        \begin{tabular}{p{15mm}p{45mm}}
        ranger: &  $mtry:\sqrt{po}$, $sample.fraction:1$, $min.node.size:1$\\
        \end{tabular}\\
    \\
    \hline
    SVM & 
        \begin{tabular}{p{18mm}p{42mm}}
            e1071: &  $C:1$, $\gamma: \frac{1}{po}$  \\
            sklearn: & $C:1$, $\gamma: \frac{1}{p * xvar}$ \\
        \end{tabular}\\ \\
    \hline
    Approx. kNN &
            \begin{tabular}{p{15mm}p{45mm}}
            mlr: &  $k:10$, $M:16$, $ef:10$, $efc:200$
            \end{tabular}\\
    \hline
    Gradient Boosting  &
            \begin{tabular}{p{15mm}p{45mm}}
            xgboost: &  $\eta:0.1$, $\lambda:1$, $\gamma:0$, $\alpha:0$, $subsample:1$,
                        $max\_depth:3$, $min\_child\_weight:1$, $colsample\_bytree:1$, $colsample\_bylevel:1$\\
            \end{tabular}\\
    \bottomrule
    \end{tabular}
    \caption{Baseline $b)$: Existing defaults for algorithm implementations. Fixed parameters described in Table \ref{tab:hpars} apply}
    \label{tab:baseline_defaults}
\end{table}

Table \ref{tab:baseline_defaults} contains existing implementation defaults used in our experiments. They have been obtained from the current versions of the implementations. We analyze algorithms from the following algorithm implementations: Elastic Net: \texttt{glmnet} \cite{glmnet} ,
Decision Trees: \texttt{rpart} \cite{rpart}, Random Forest: \texttt{ranger} \cite{ranger}, SVM: \texttt{LibSVM} via \texttt{e1071}  (\cite{libsvm}, \cite{e1071}) and \texttt{xgboost} \cite{xgboost}. We investigate \texttt{HNSW} \cite{hnsw} as an approximate k-Nearest-Neighbours algorithm. Additional details on the exact meaning of the different hyperparameters can be obtained from the respective software's documentation. We assume that small differences due to implementation details e.g. between the \texttt{LibSVM} and \texttt{sklearn} implementations exist, but try to compare to existing default settings nonetheless, as they might serve as relevant baselines.

\section{Evolutionary Search}
\label{app:mpl}

The main components of the evolutionary search are given in Section~\ref{sec:alg}.
This appendix only defines the mutation operations.
We use a number of mutation operators, but not all mutations can be applied to all candidates.
Given a parent, we first determine the mutations that will lead to valid offspring, after which we apply one chosen uniformly at random.
The mutation operators are:\\
\noindent\textbf{Node Insertion}: Pick a node in the tree and add an operator node between that node and its parent node. If necessary, additional input to the new node is generated by randomly selecting terminals. \\
\textbf{Shrink Node}: Select a node and replace it with one of its subtrees.\\
\textbf{Node Replacement}: Replace a randomly chosen node by another node of the same type and arity. \\
\textbf{Terminal Replacement}: Replace a terminal with a different terminal of the same type (i.e. <I> or <F>).\\
\textbf{Mutate Ephemeral}: Change the value of an ephemeral constant (i.e. $c_i$ or $c_f$) with Gaussian noise proportional to the ephemeral's value. For the integer ephemeral, the change is rounded and can not be zero.\\
None of the mutations that work on operators work on the <configuration> operator.\\
In order to define the search space for symbolic formulas, we define a grammar composed of terminal symbols and operators. 
Operators can take one or multiple terminals or operators as input and produce a single output. 
We consider two types of terminal symbols: ephemeral constants and meta-features.
This allows for a very flexible description of the search space.

\subsubsection*{$\mu + \lambda$ vs. random search}
\label{app:rq1}
\begin{figure}[h!]
    \centering
    \includegraphics[width=0.4\textwidth]{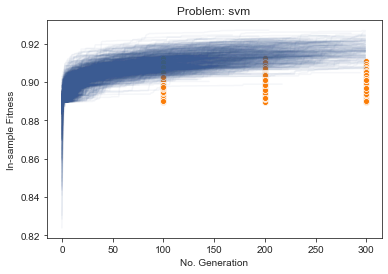}
    \caption{In-sample fitness scores of $\mu + \lambda$ (blue) and $100$, $200$ and $300$ generations equivalent of random search (orange).}
    \label{fig:opt_trace}
\end{figure}

Figure \ref{fig:opt_trace} depicts optimization traces of $\mu + \lambda$ and random search across $10$ replications on all datasets. Shorter EA traces occur due to early stopping. Genetic programming seems to consistently yield better results.

\section{Experimental Results}
\label{app:exps_surr}
The following section describes the results of the Experiments conducted to answer \textbf{RQ1} and \textbf{RQ2} across all other algorithms analyzed in this paper. Results and a more detailed analysis for the SVM can be obtained from section \ref{sec:exps_rq2}. 

\newpage

\subsection{Elastic Net}
\vspace{-.2cm}
\begin{figure}[h]
    \begin{subfigure}[c]{0.45\textwidth}
        \centering
        \includegraphics[width=\textwidth]{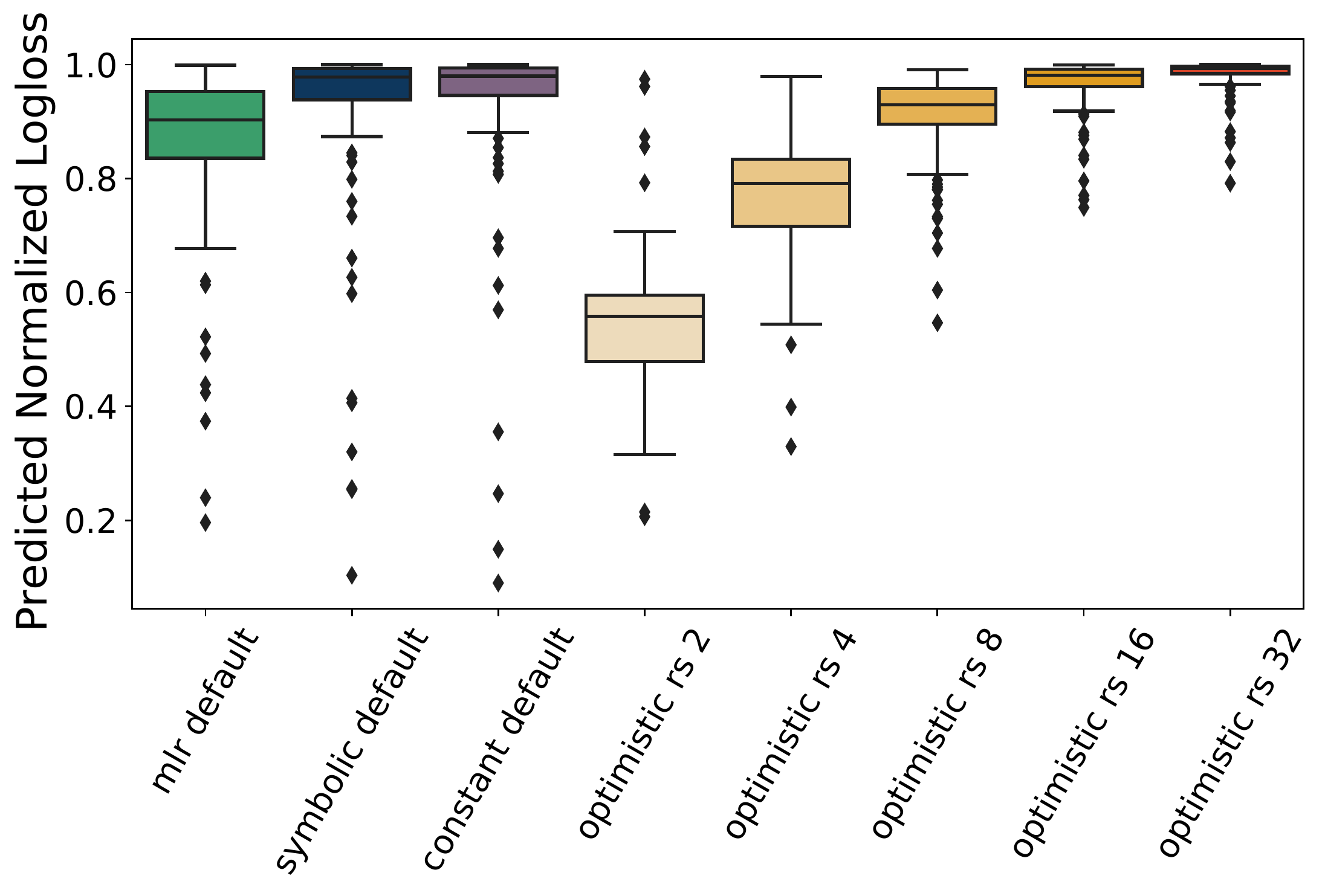}
        \caption{Symbolic, static and implementation defaults, comparing scaled logloss predicted by surrogates.}
    \end{subfigure}
    \begin{subfigure}[c]{0.5\textwidth}
        \centering
        \includegraphics[width=\textwidth]{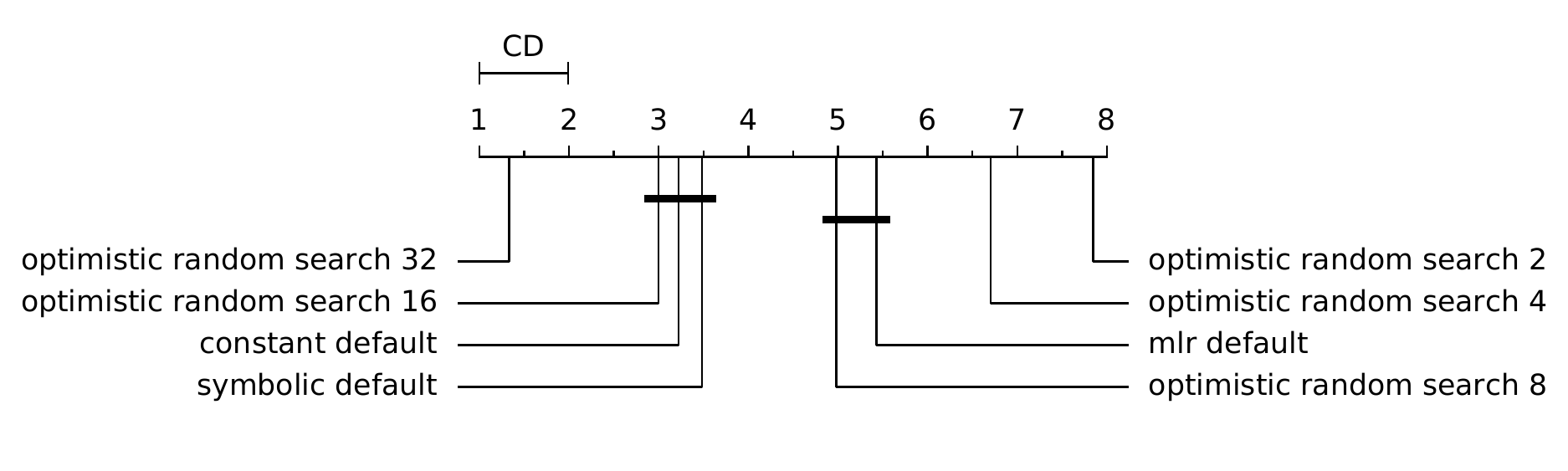}
        \caption{Critical Differences Diagram of symbolic, static and implementation defaults on surrogates}
    \end{subfigure}
    \begin{subfigure}[c]{0.5\textwidth}
        \centering
        \begin{subfigure}[c]{0.48\textwidth}
        \includegraphics[width=\textwidth]{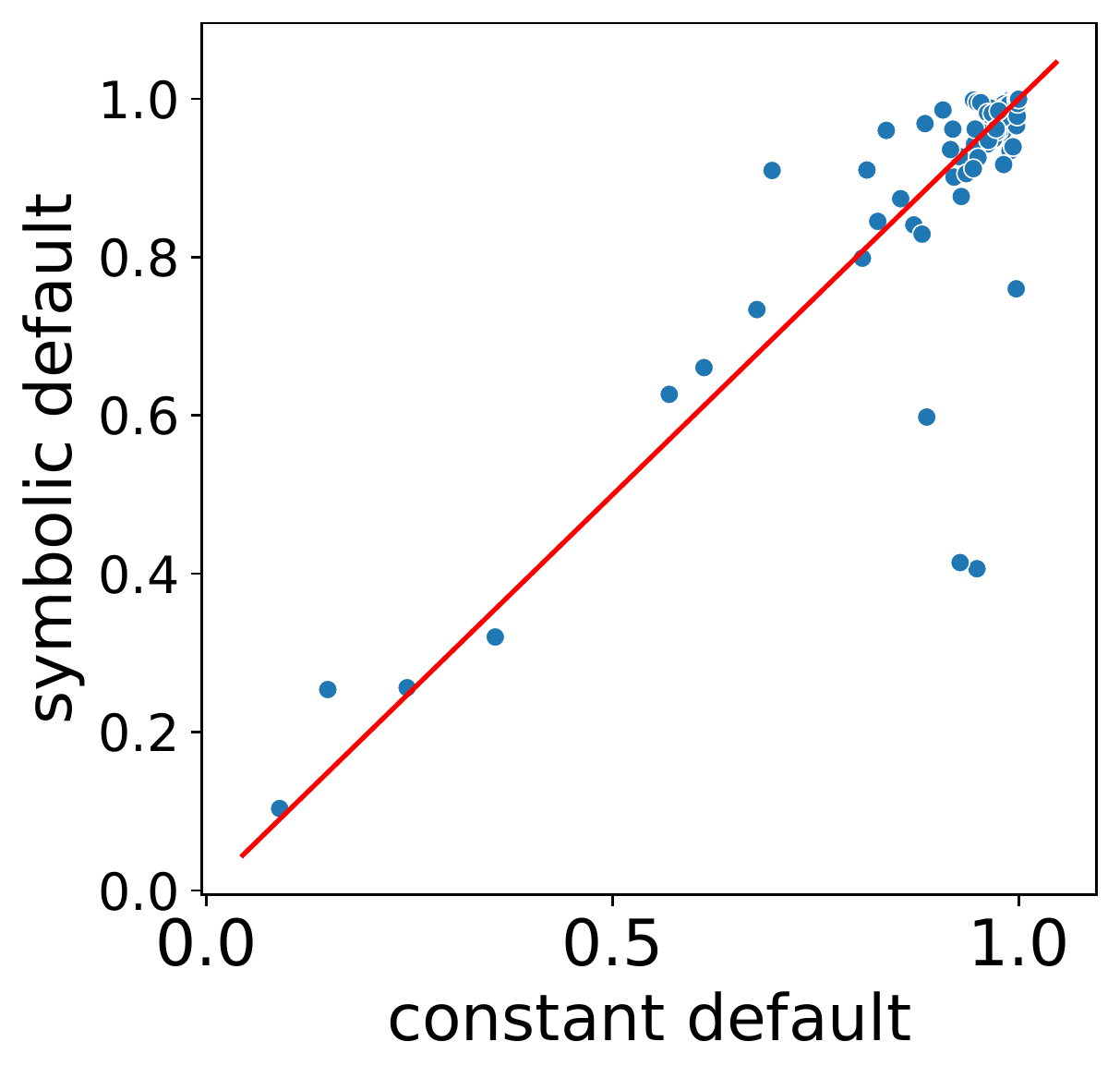}
        \end{subfigure}
        \begin{subfigure}[c]{0.48\textwidth}
        \includegraphics[width=\textwidth]{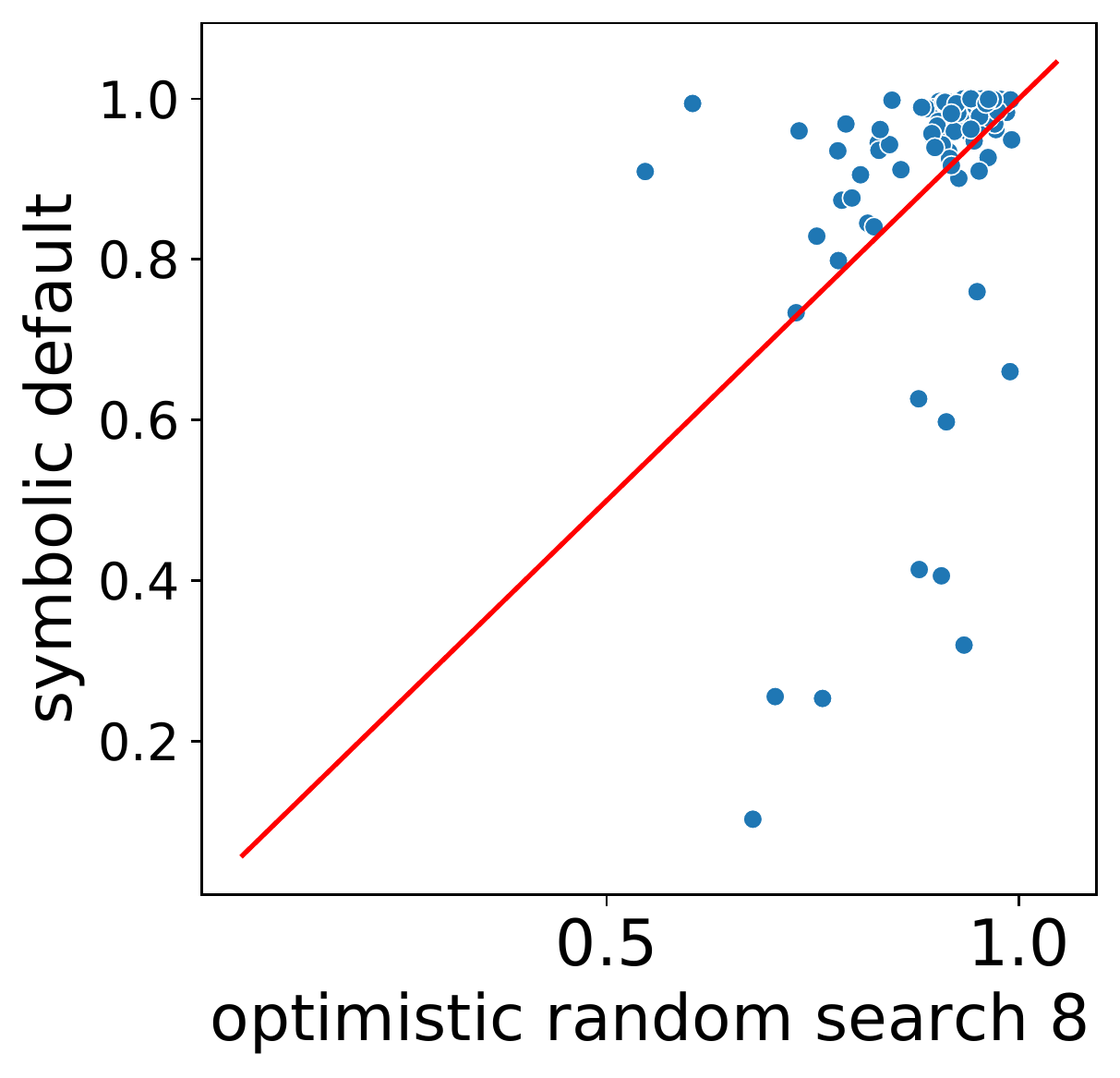}
        \end{subfigure}
        \caption{Performance comparison of symbolic defaults to constant defaults (left) and budget 8 optimistic random search (right).}
    \end{subfigure}
    \caption{Results for the elastic net algorithm on surrogate data.}
    \label{fig:results_glmnet}
\end{figure}

\newpage

\subsection{Decision Trees}
\vspace{-.2cm}
\begin{figure}[h]
    \begin{subfigure}[c]{0.45\textwidth}
        \centering
        \includegraphics[width=\textwidth]{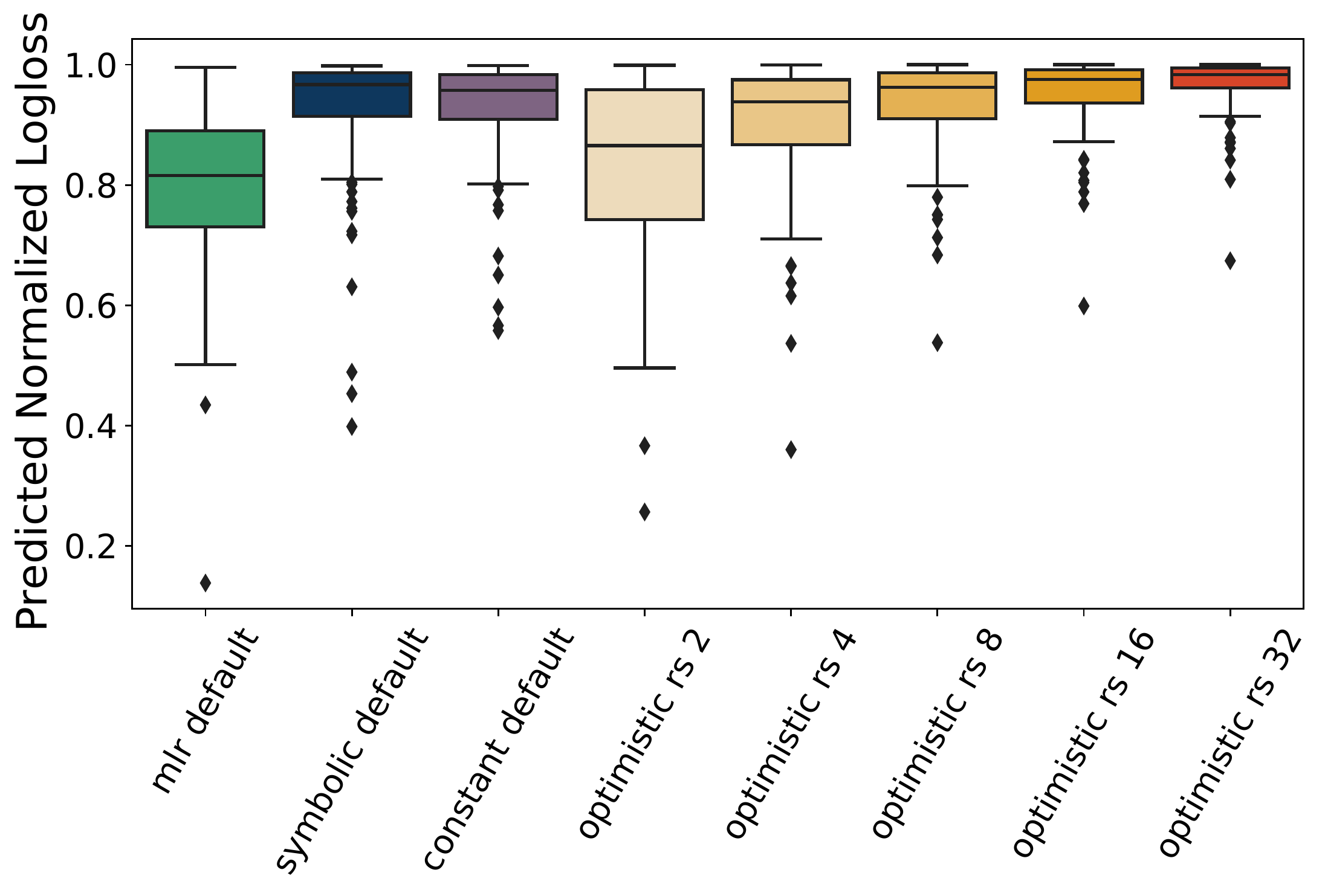}
        \caption{Symbolic, static and implementation defaults, comparing scaled logloss predicted by surrogates.}
    \end{subfigure}
    \begin{subfigure}[c]{0.5\textwidth}
        \centering
        \includegraphics[width=\textwidth]{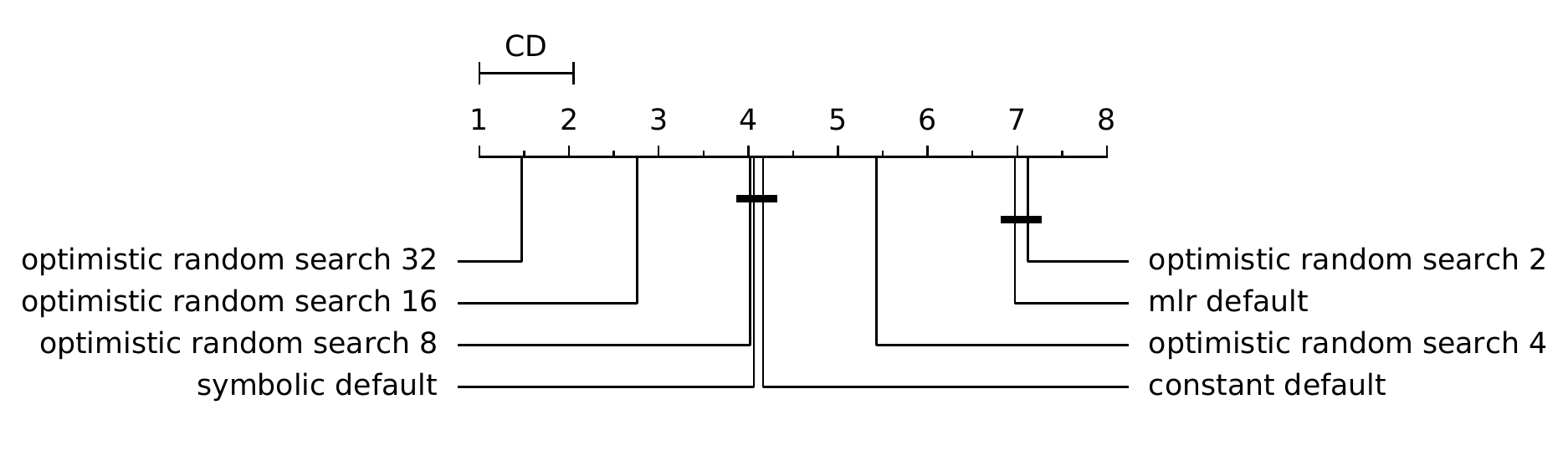}
        \caption{Critical Differences Diagram of symbolic, static and implementation defaults on surrogates}
    \end{subfigure}
    \begin{subfigure}[c]{0.5\textwidth}
        \centering
        \begin{subfigure}[c]{0.48\textwidth}
        \includegraphics[width=\textwidth]{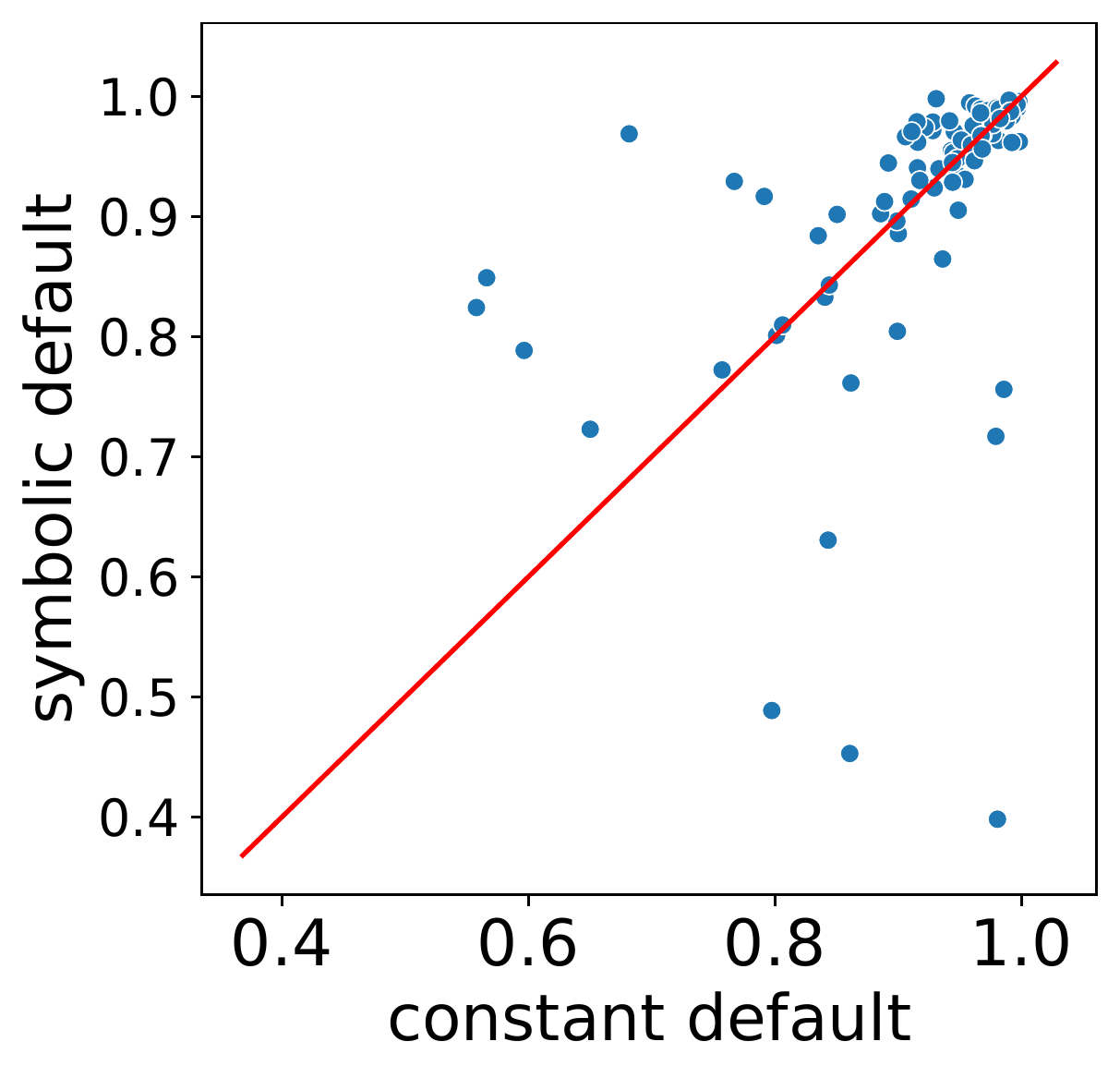}
        \end{subfigure}
        \begin{subfigure}[c]{0.48\textwidth}
        \includegraphics[width=\textwidth]{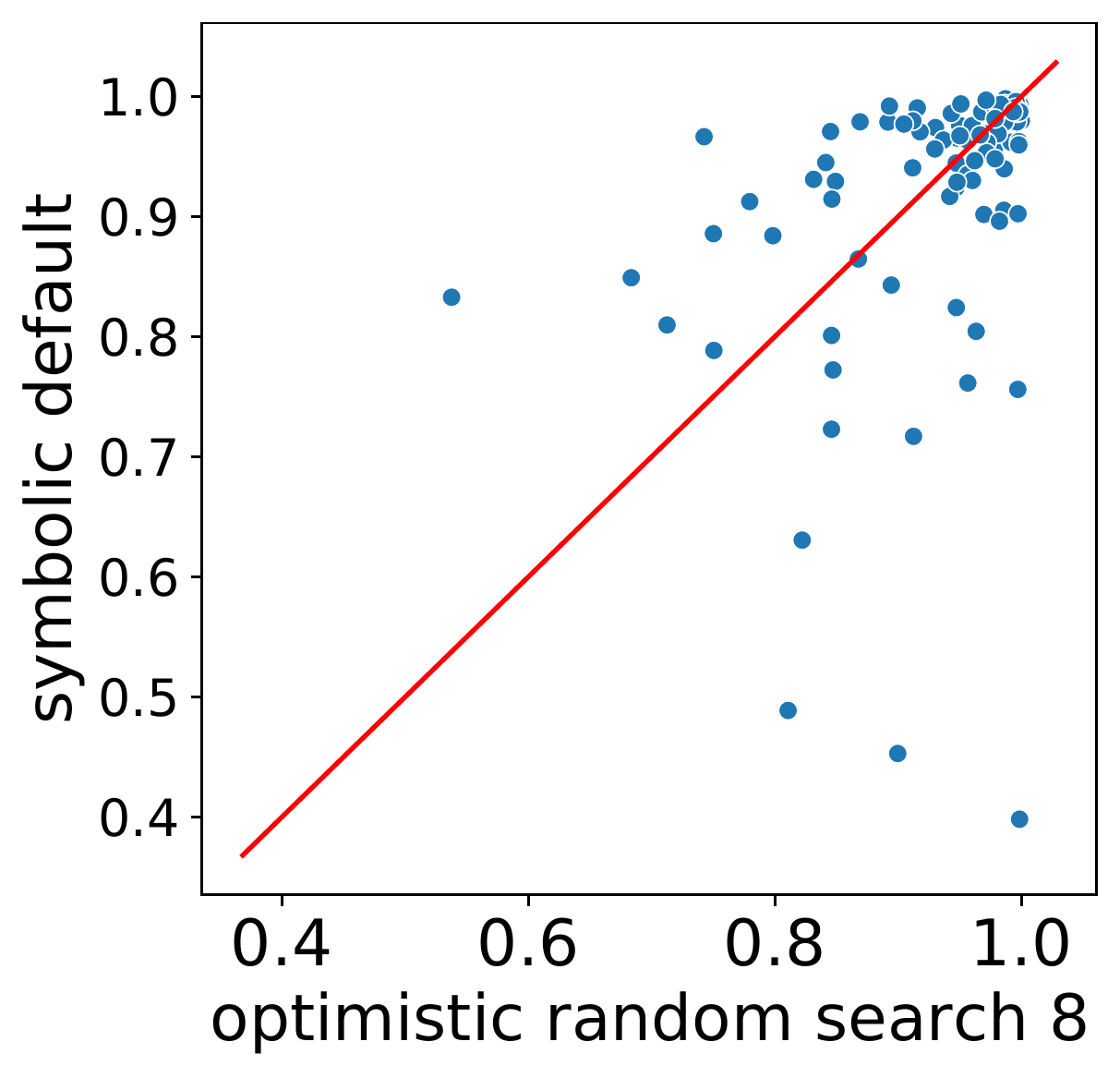}
        \end{subfigure}
        \caption{Performance comparison of symbolic defaults to constant defaults (left) and budget 8 optimistic random search (right).}
    \end{subfigure}
    \caption{Results for the decision tree algorithm on surrogate data.}
    \label{fig:results_rpart}
\end{figure}

\newpage

\subsection{Approximate k-Nearest Neighbours}
\vspace{-.2cm}
\begin{figure}[h]
    \begin{subfigure}[c]{0.45\textwidth}
        \centering
        \includegraphics[width=\textwidth]{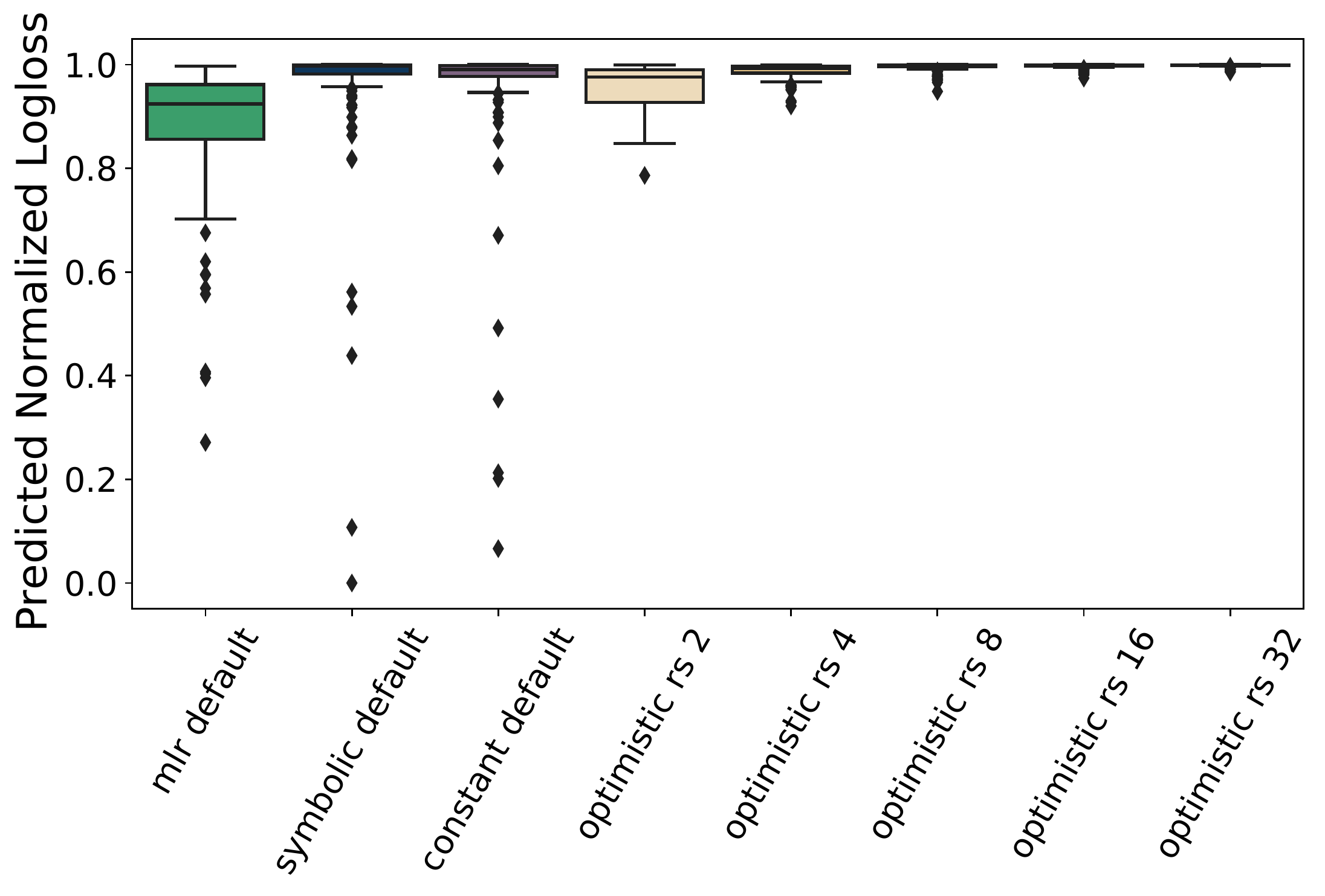}
        \caption{Symbolic, static and implementation defaults, comparing scaled logloss predicted by surrogates.}
    \end{subfigure}
    \begin{subfigure}[c]{0.5\textwidth}
        \centering
        \includegraphics[width=\textwidth]{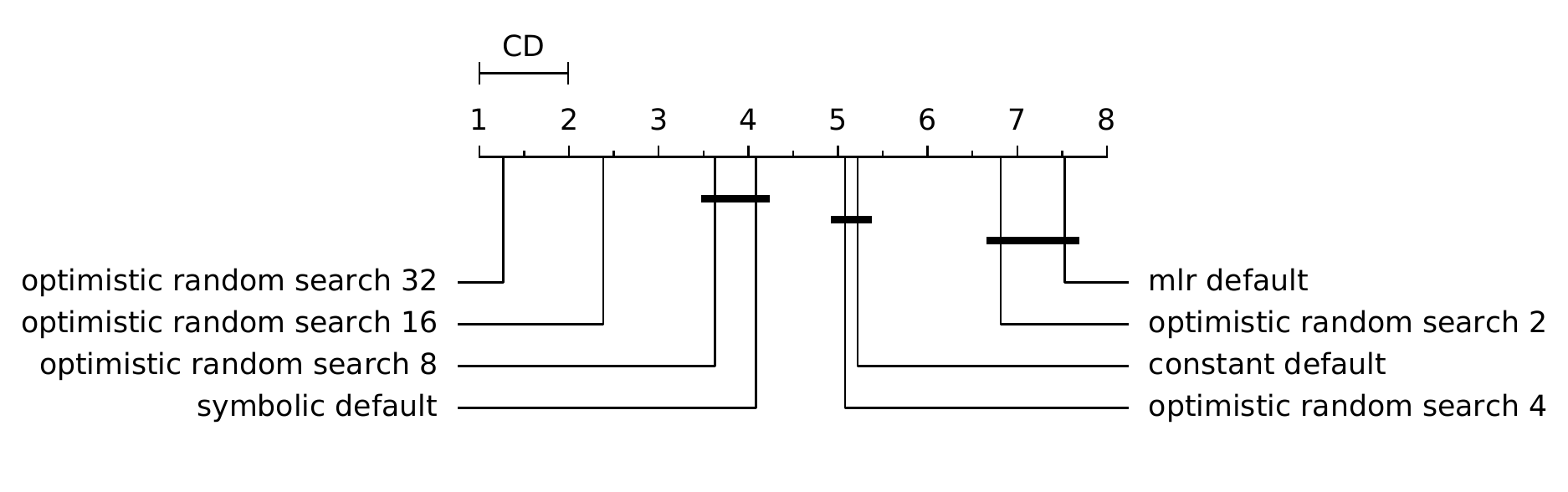}
        \caption{Critical Differences Diagram of symbolic, static and implementation defaults on surrogates}
    \end{subfigure}
    \begin{subfigure}[c]{0.5\textwidth}
        \centering
        \begin{subfigure}[c]{0.48\textwidth}
        \includegraphics[width=\textwidth]{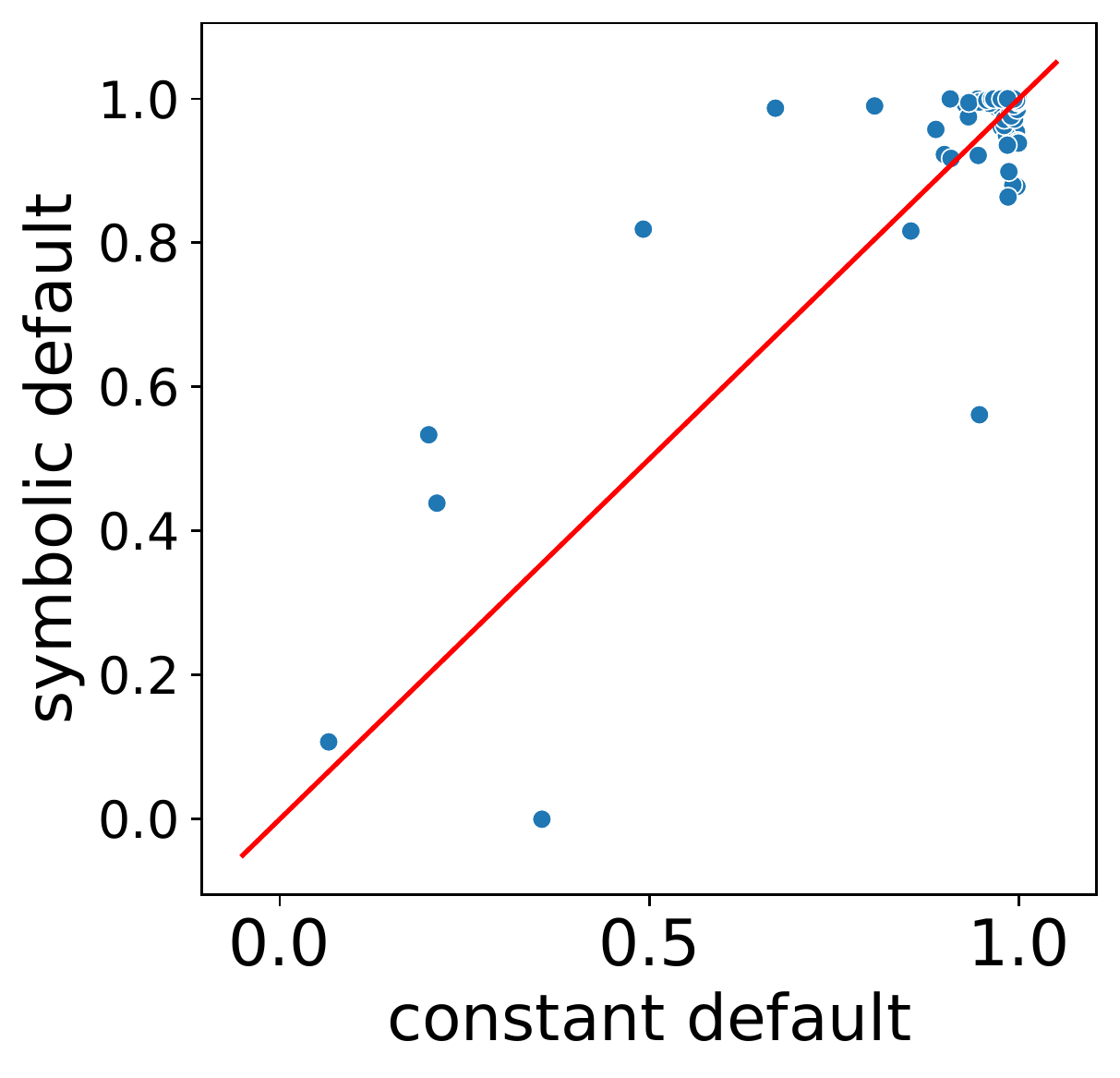}
        \end{subfigure}
        \begin{subfigure}[c]{0.48\textwidth}
        \includegraphics[width=\textwidth]{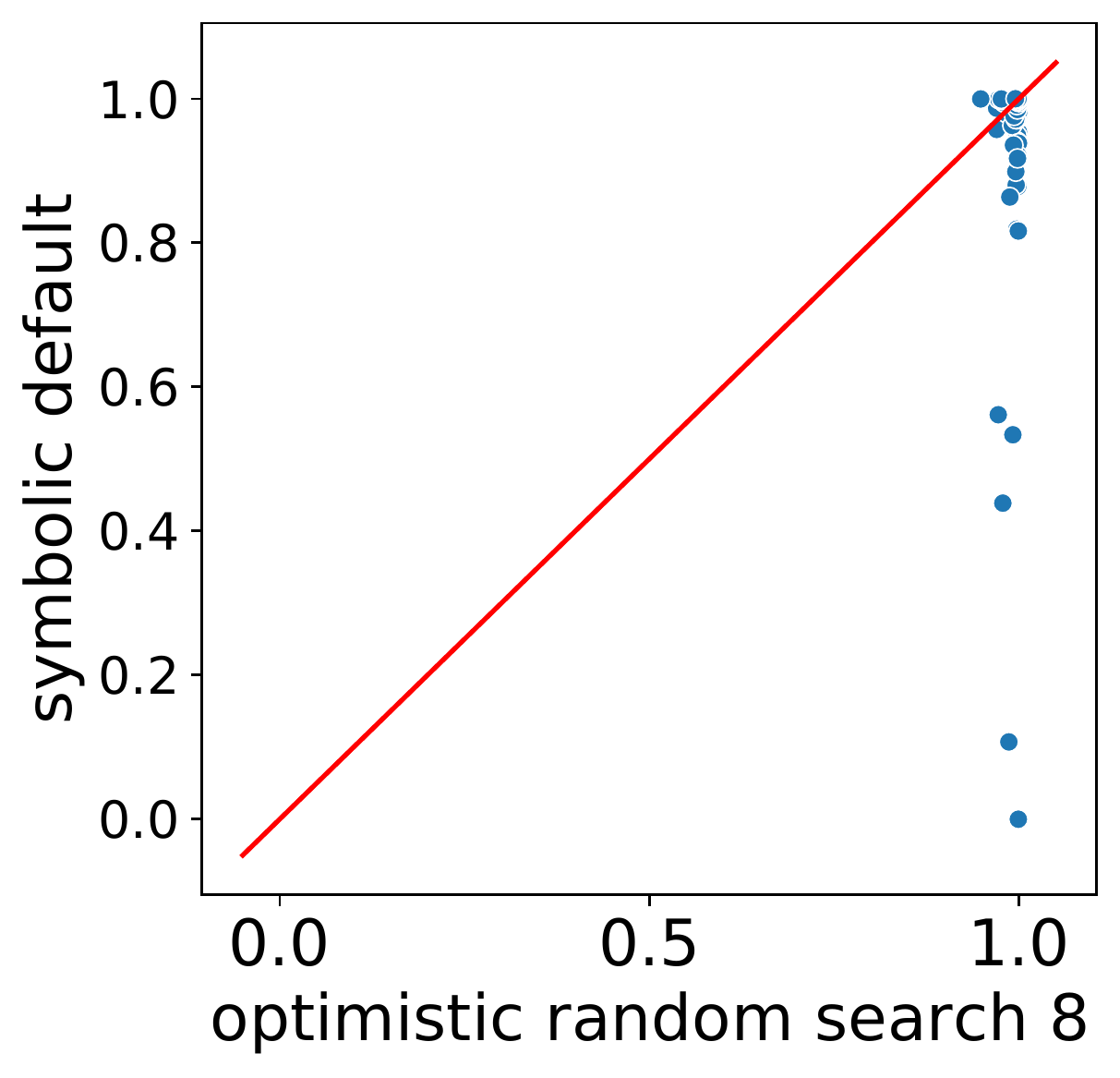}
        \end{subfigure}
        \caption{Performance comparison of symbolic defaults to constant defaults (left) and budget 8 optimistic random search (right).}
    \end{subfigure}
    \caption{Results for the approximate k-nearest neighbours algorithm on surrogate data.}
    \label{fig:results_knn}
\end{figure}

\newpage

\subsection{Random Forest}
\vspace{-.2cm}
\begin{figure}[h]
    \begin{subfigure}[c]{0.45\textwidth}
        \centering
        \includegraphics[width=\textwidth]{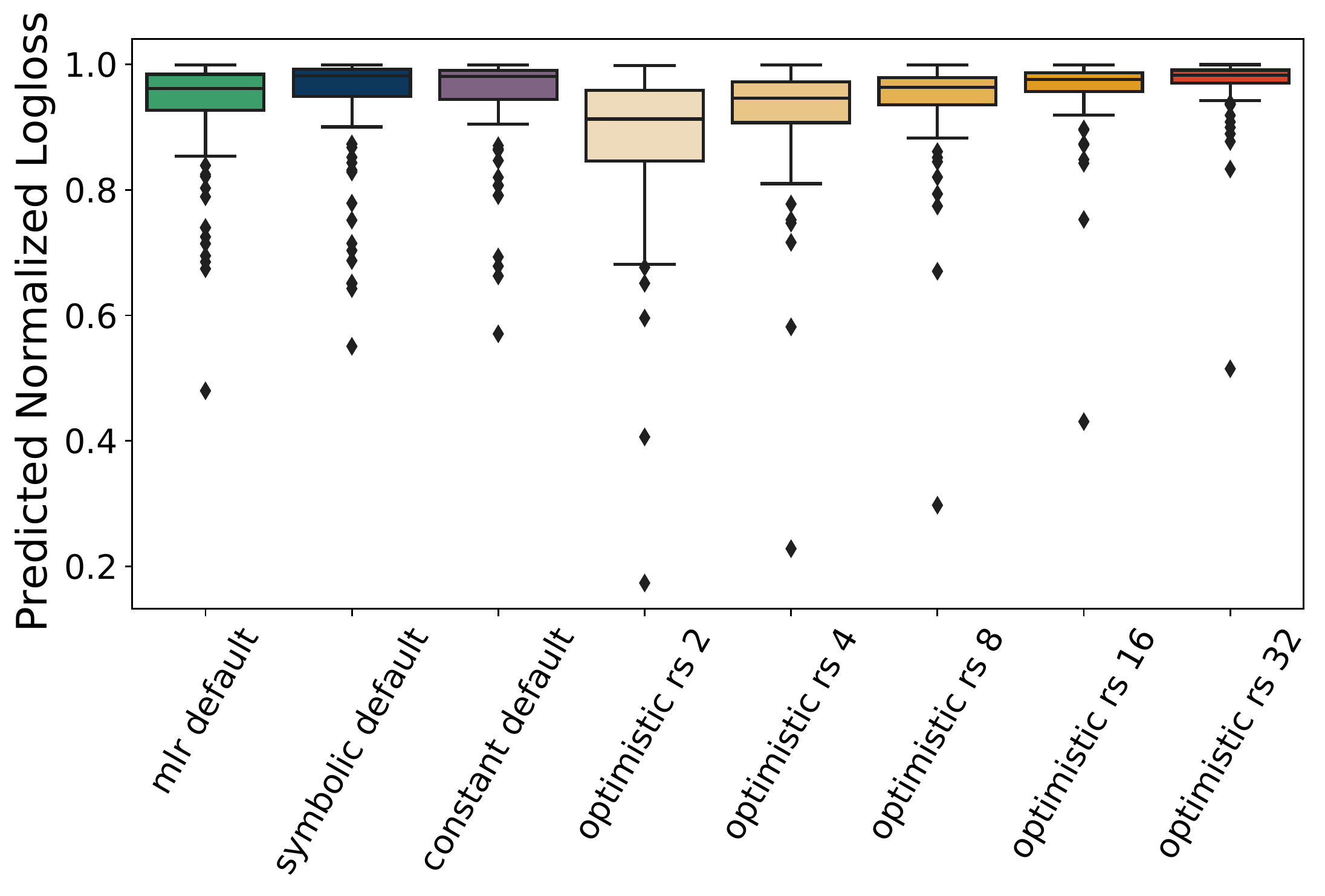}
        \caption{Symbolic, static and implementation defaults, comparing scaled logloss predicted by surrogates.}
    \end{subfigure}
    \begin{subfigure}[c]{0.5\textwidth}
        \centering
        \includegraphics[width=\textwidth]{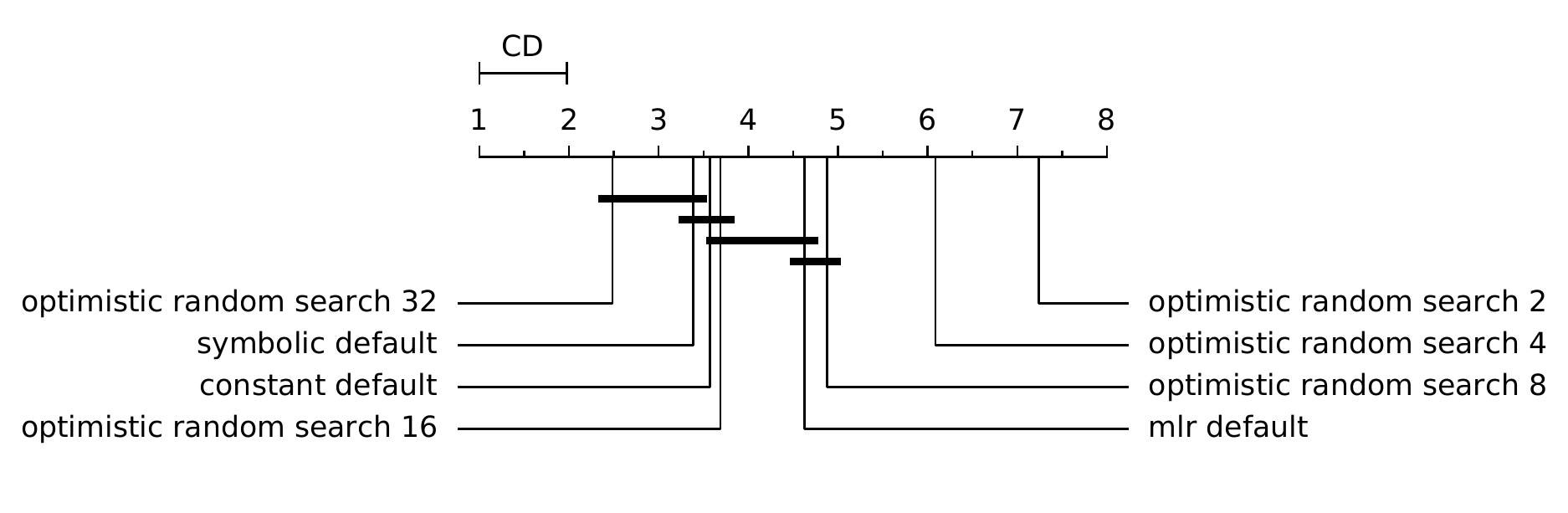}
        \caption{Critical Differences Diagram of symbolic, static and implementation defaults on surrogates}
    \end{subfigure}
    \begin{subfigure}[c]{0.5\textwidth}
        \centering
        \begin{subfigure}[c]{0.48\textwidth}
        \includegraphics[width=\textwidth]{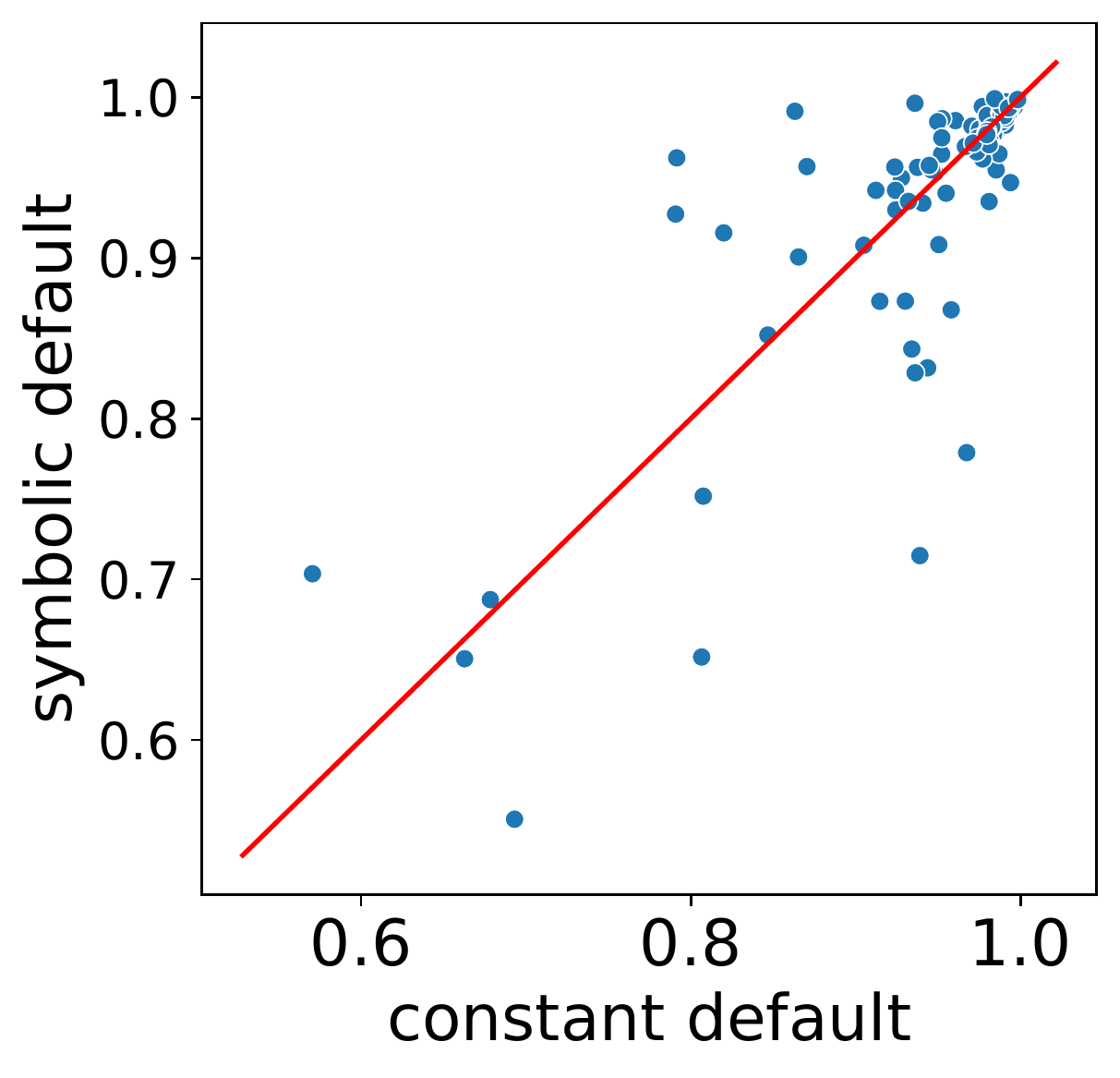}
        \end{subfigure}
        \begin{subfigure}[c]{0.48\textwidth}
        \includegraphics[width=\textwidth]{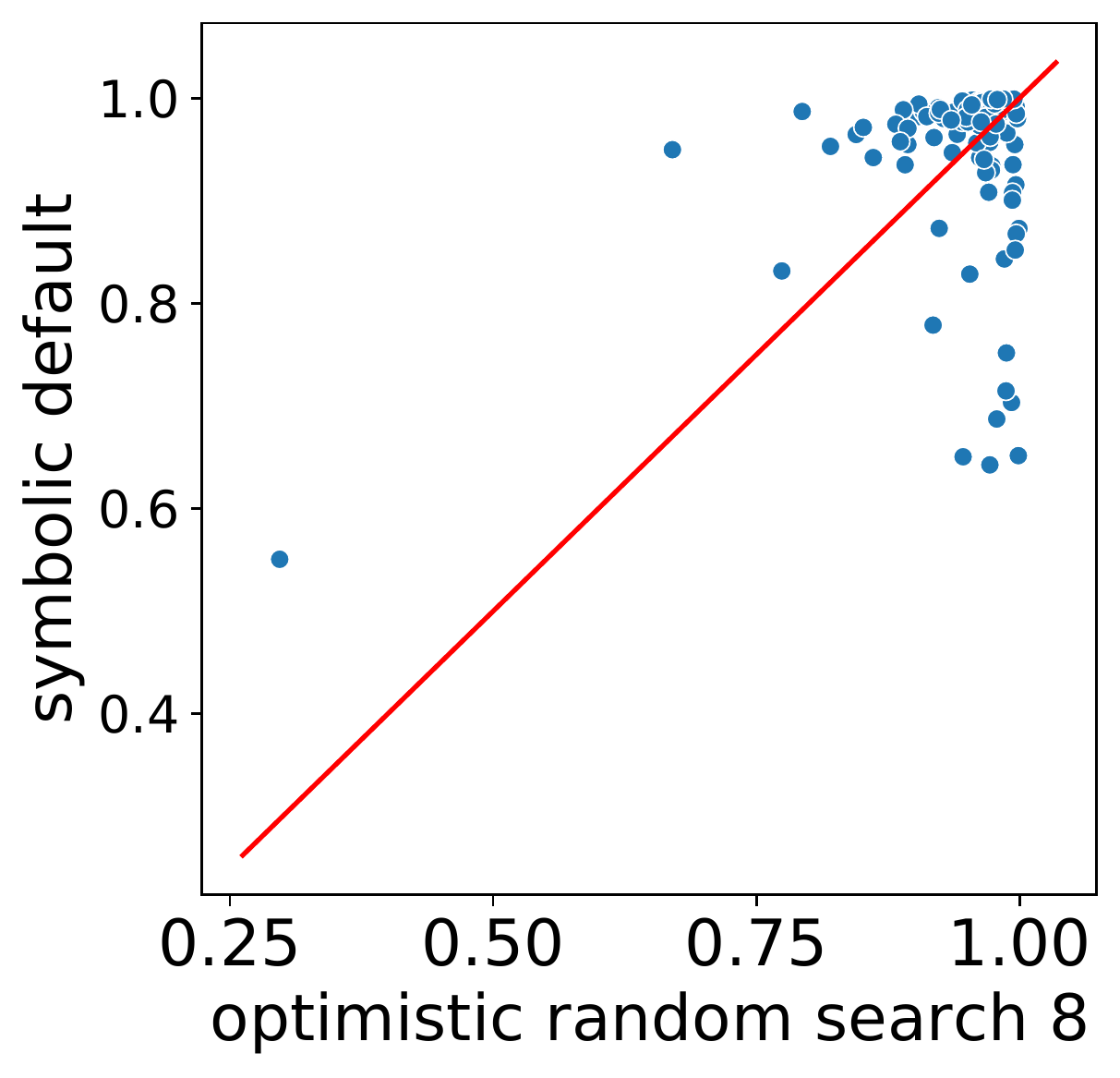}
        \end{subfigure}
        \caption{Performance comparison of symbolic defaults to constant defaults (left) and budget 8 optimistic random search (right).}
    \end{subfigure}
    \caption{Results for the random forest algorithm on surrogate data.}
    \label{fig:results_rf}
\end{figure}

\newpage

\subsection{eXtreme Gradient Boosting (XGBoost)}
\vspace{-.2cm}
\begin{figure}[h]
    \begin{subfigure}[c]{0.45\textwidth}
        \centering
        \includegraphics[width=\textwidth]{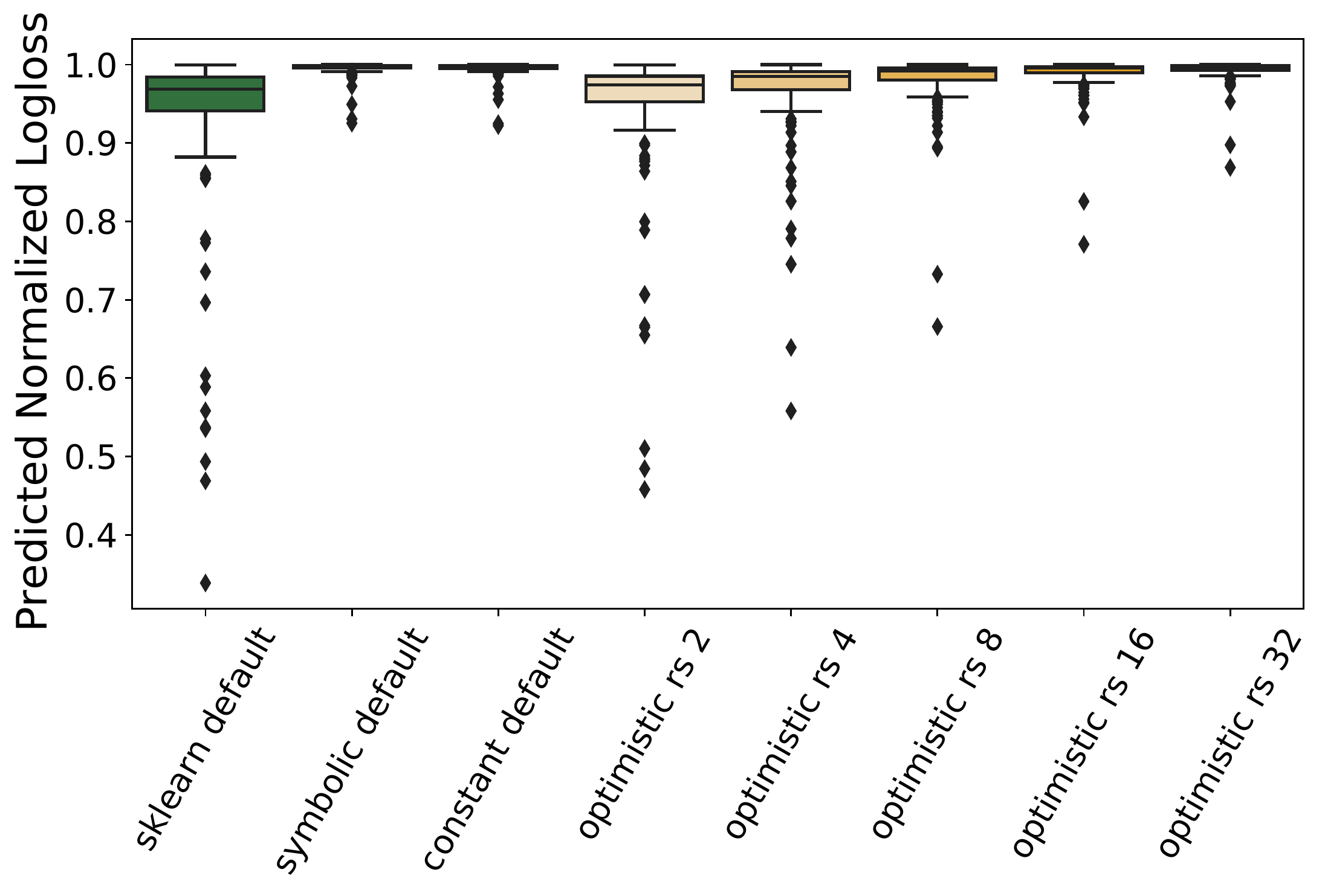}
        \caption{Symbolic, static and implementation defaults, comparing scaled logloss predicted by surrogates.}
    \end{subfigure}
    \begin{subfigure}[c]{0.5\textwidth}
        \centering
        \includegraphics[width=\textwidth]{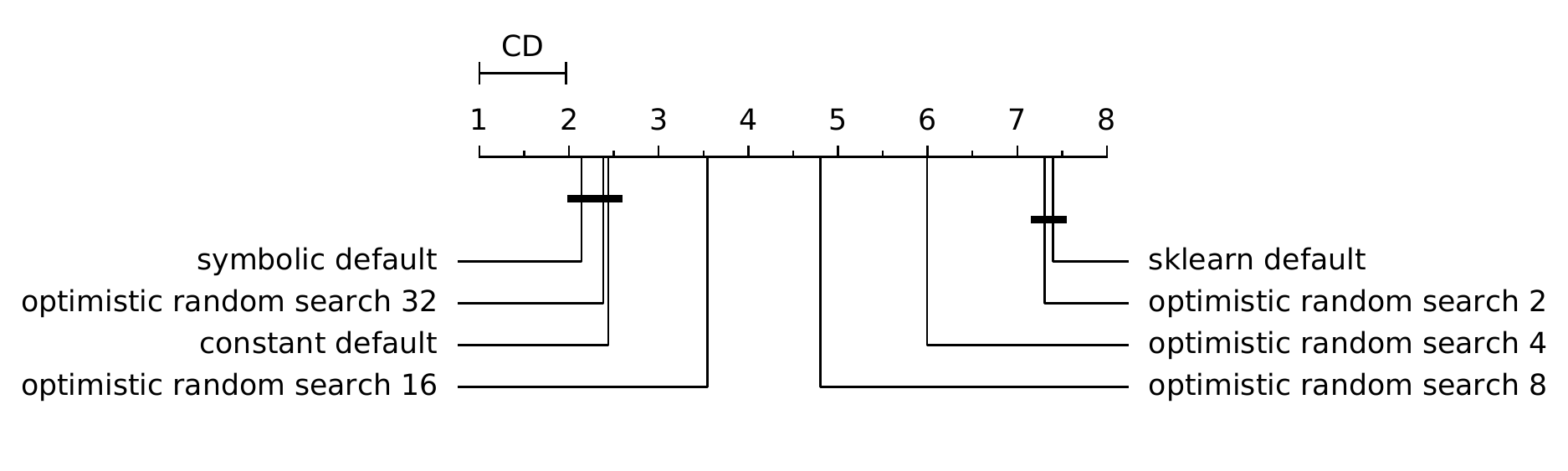}
        \caption{Critical Differences Diagram of symbolic, static and implementation defaults on surrogates}
    \end{subfigure}
    \begin{subfigure}[c]{0.5\textwidth}
        \centering
        \begin{subfigure}[c]{0.48\textwidth}
        \includegraphics[width=\textwidth]{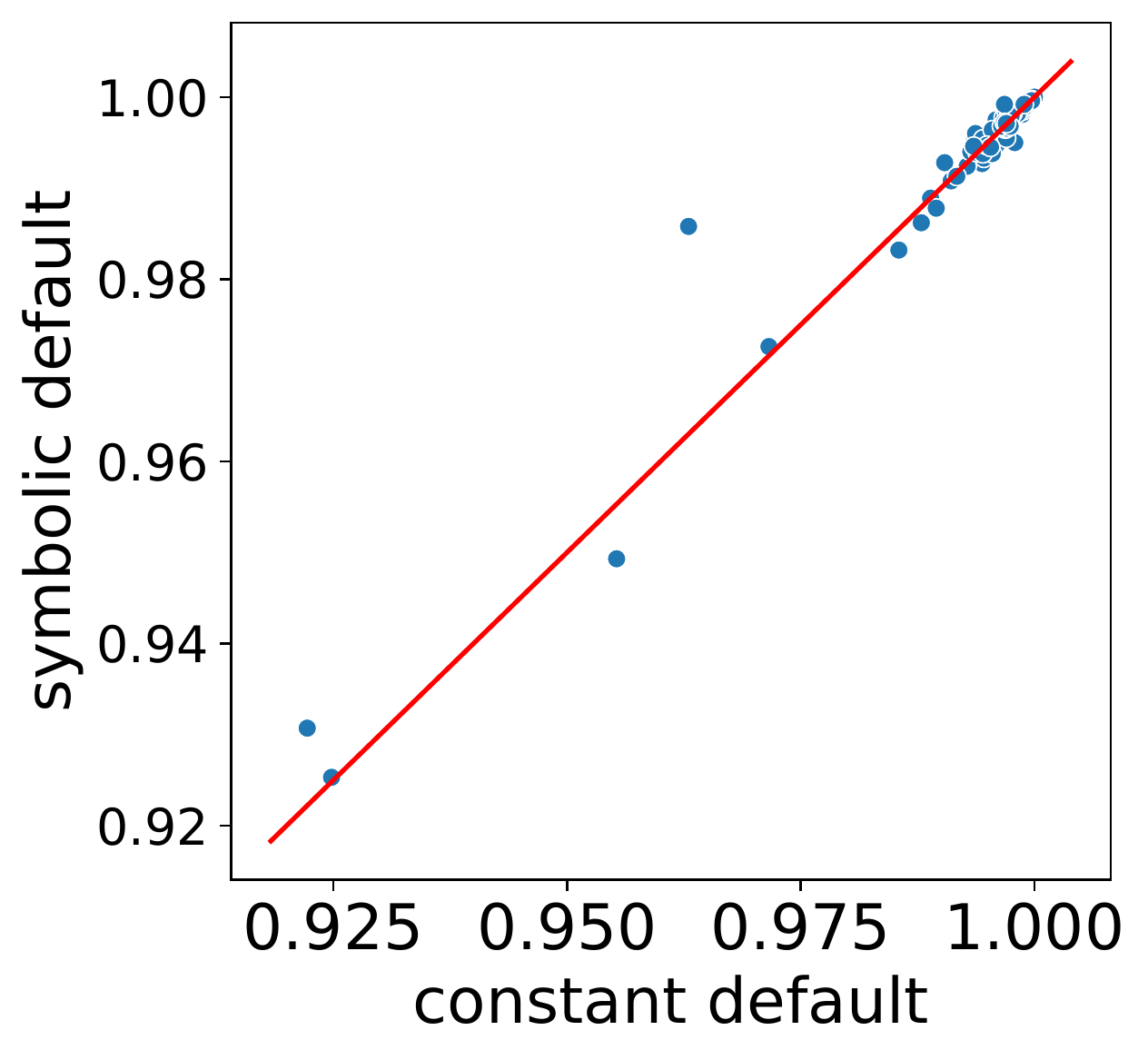}
        \end{subfigure}
        \begin{subfigure}[c]{0.48\textwidth}
        \includegraphics[width=\textwidth]{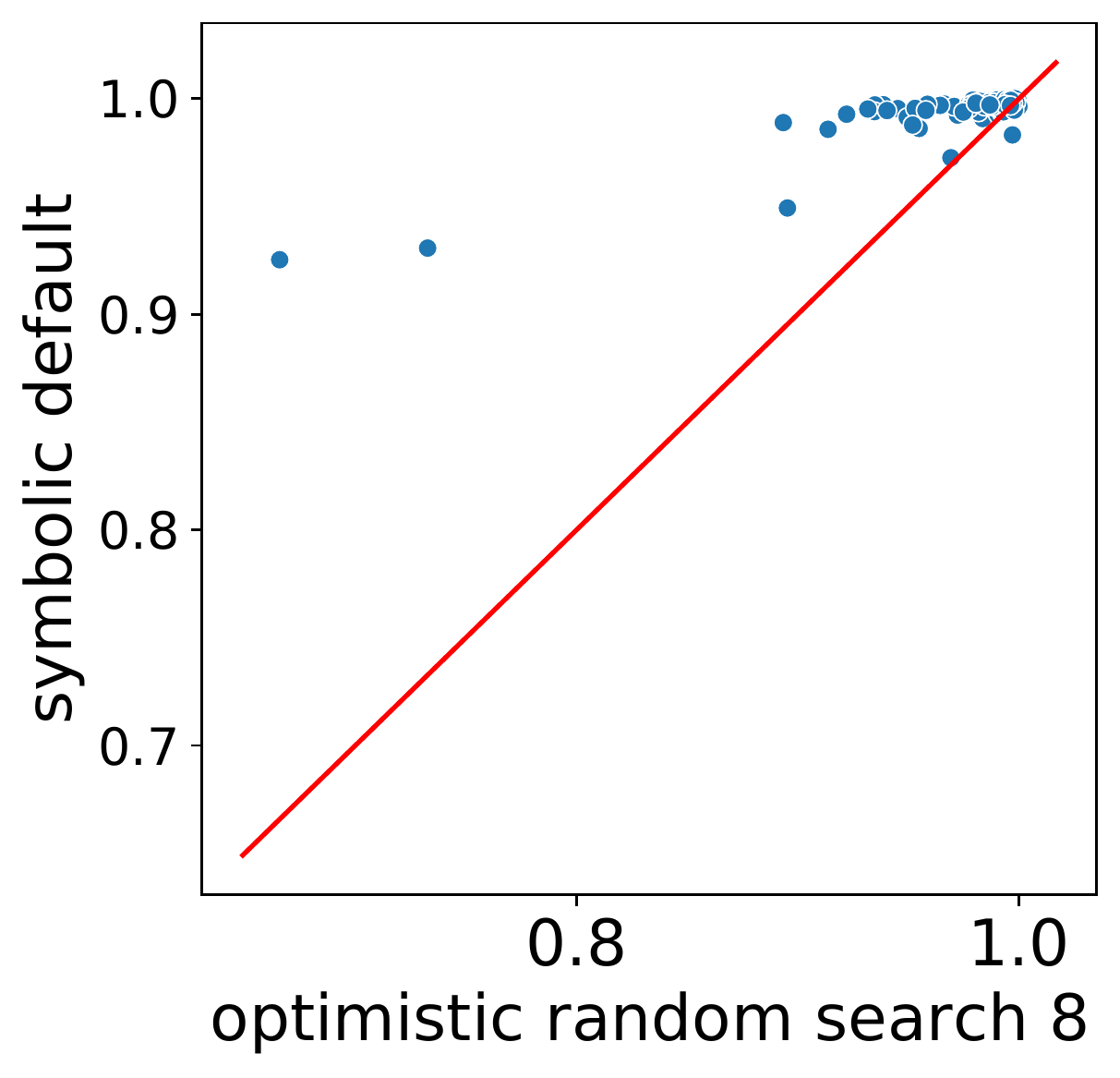}
        \end{subfigure}
        \caption{Performance comparison of symbolic defaults to constant defaults (left) and budget 8 optimistic random search (right).}
    \end{subfigure}
    \caption{Results for the XGBoost algorithm on surrogate data.}
    \label{fig:results_xgboost}
\end{figure}

\section{Real Data Experiments}

In analogy to the presentation of the results for the SVM of the main text, we present results for Decision Tree and Elastic Net here. 
\subsection{Decision Tree}

\begin{figure}[h]
    \centering
\begin{subfigure}[c]{0.24\textwidth}
    \centering
    \includegraphics[width=\textwidth]{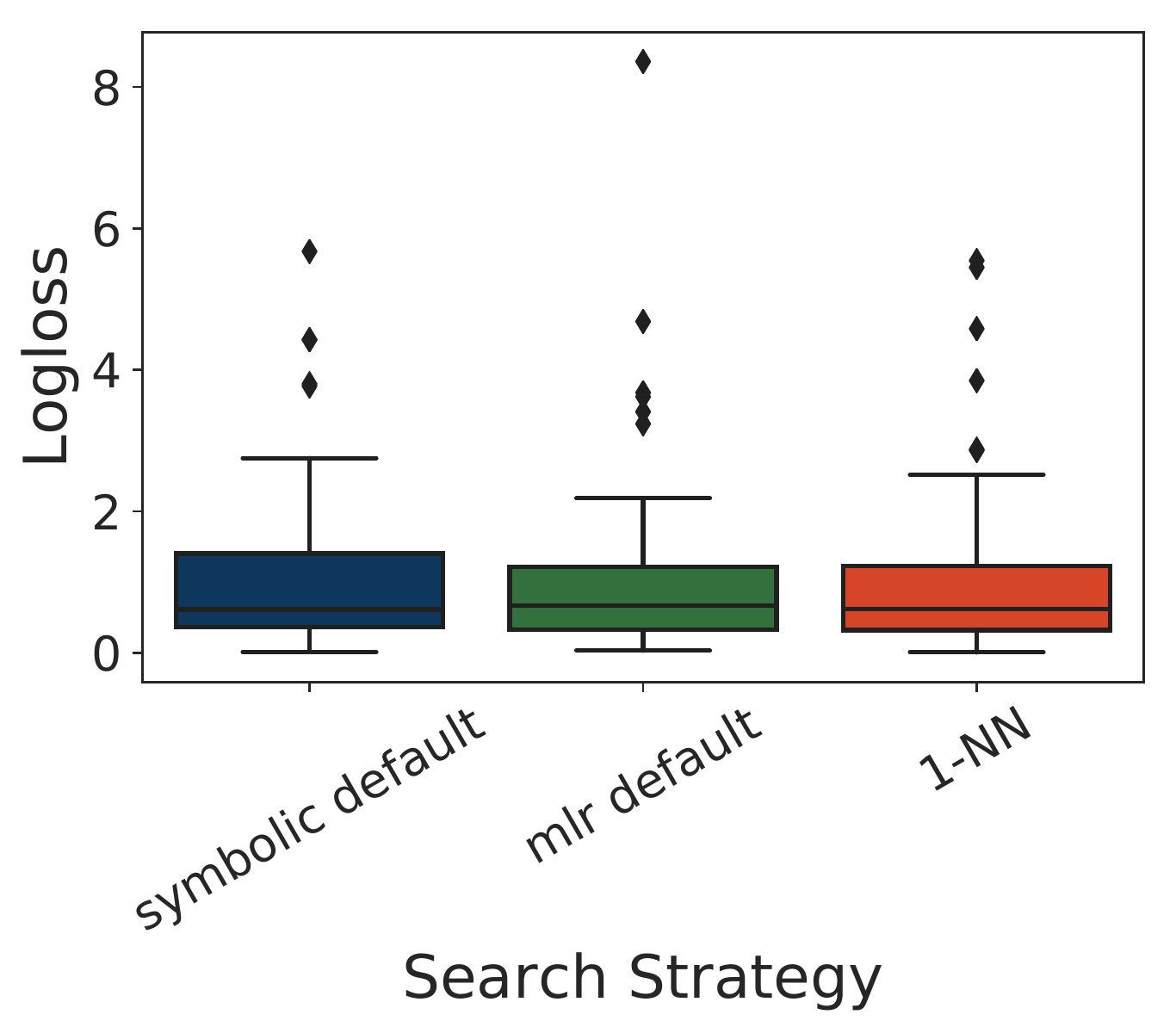}
\end{subfigure}
\begin{subfigure}[c]{0.23\textwidth}
    \centering
    \includegraphics[width=\textwidth]{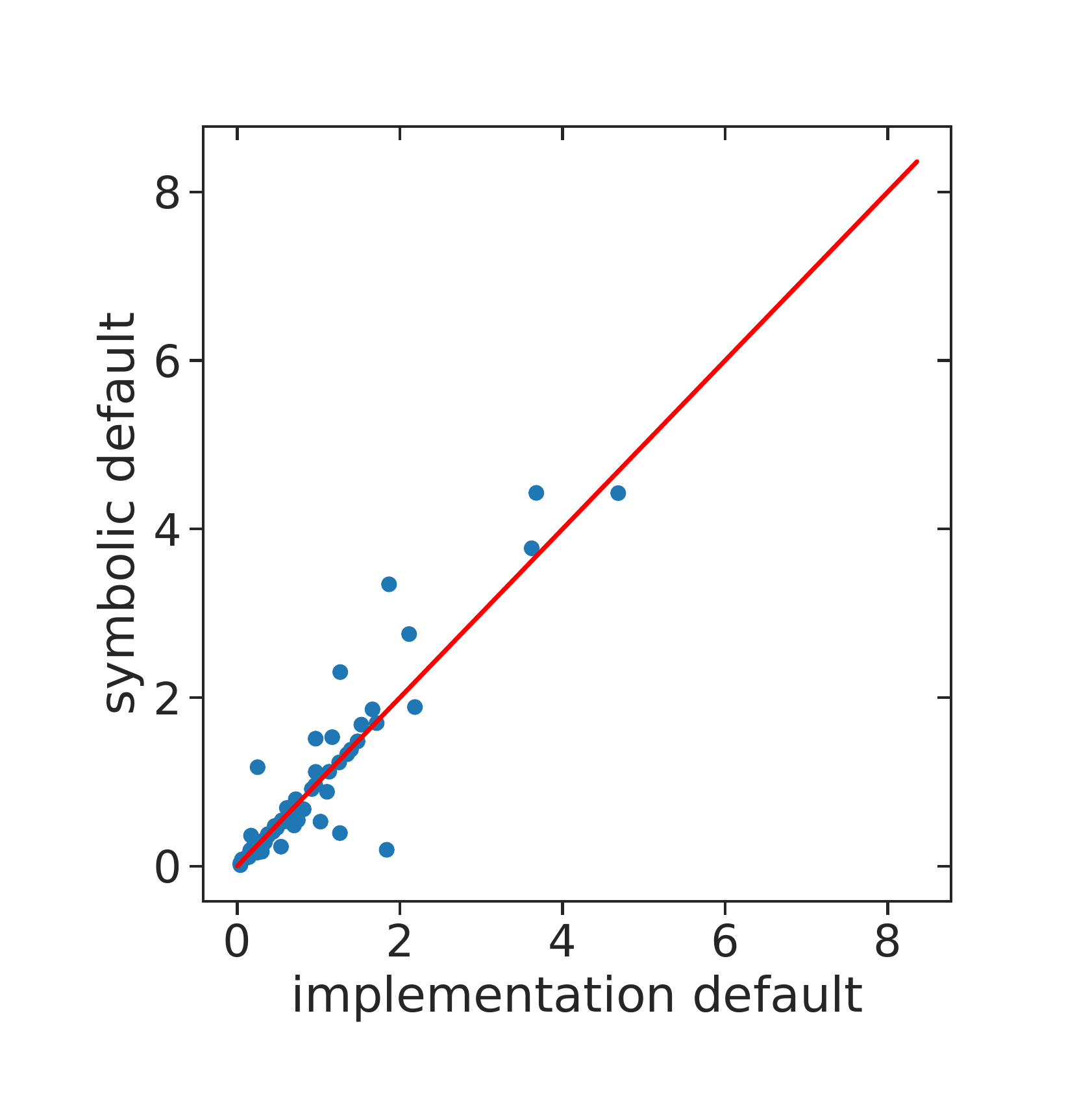}
\end{subfigure}
    \caption{Results for the decision tree algorithm. Comparison of symbolic and implementation default using log-loss across all datasets performed on real data. Box plots (right) and scatter plot (left)}
    \label{fig:svm_realdata}
\end{figure}
\FloatBarrier
\subsection{Elastic Net}

\begin{figure}[h]
    \centering
\begin{subfigure}[c]{0.24\textwidth}
    \centering
    \includegraphics[width=\textwidth]{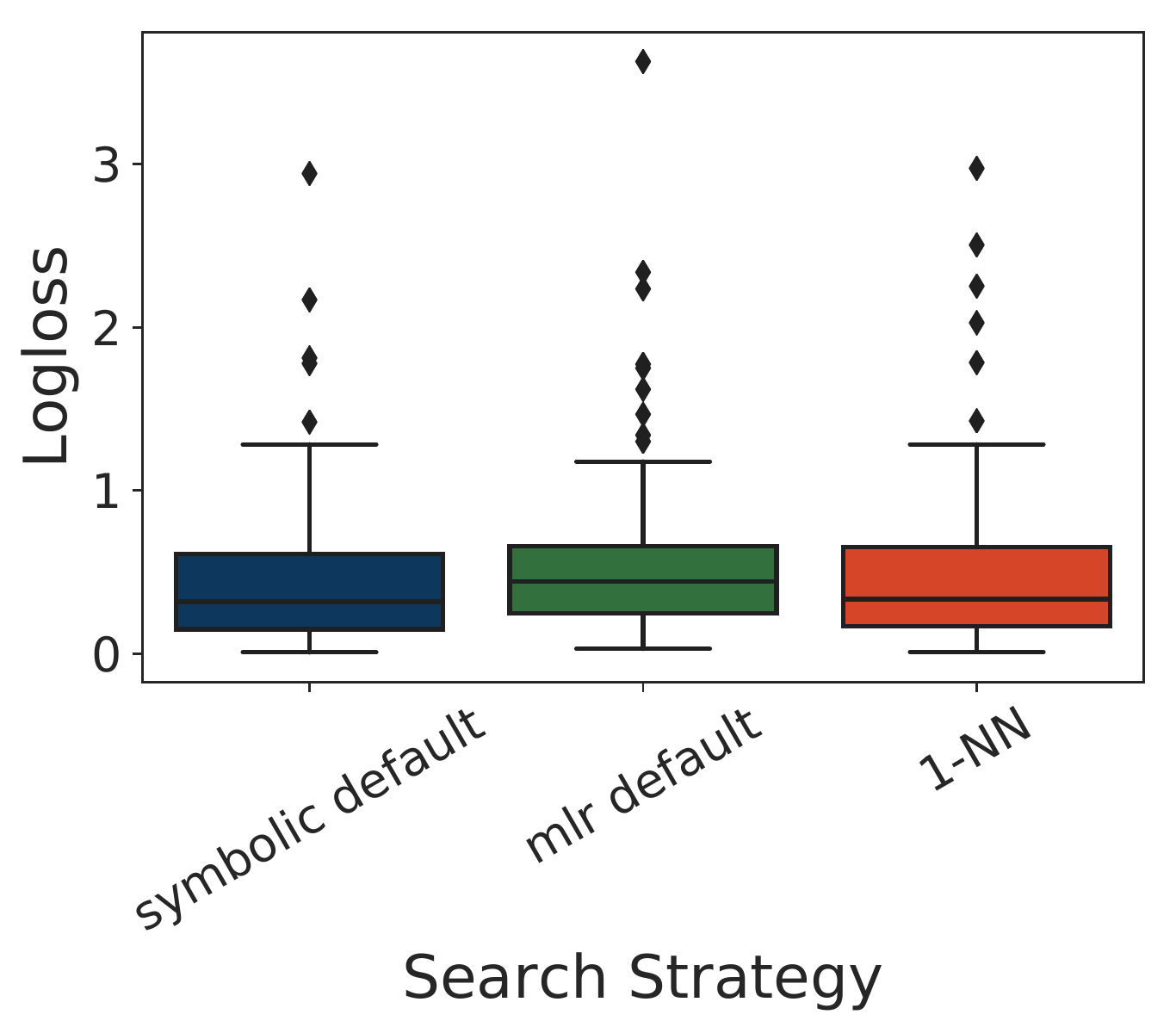}
\end{subfigure}
\begin{subfigure}[c]{0.23\textwidth}
    \centering
    \includegraphics[width=\textwidth]{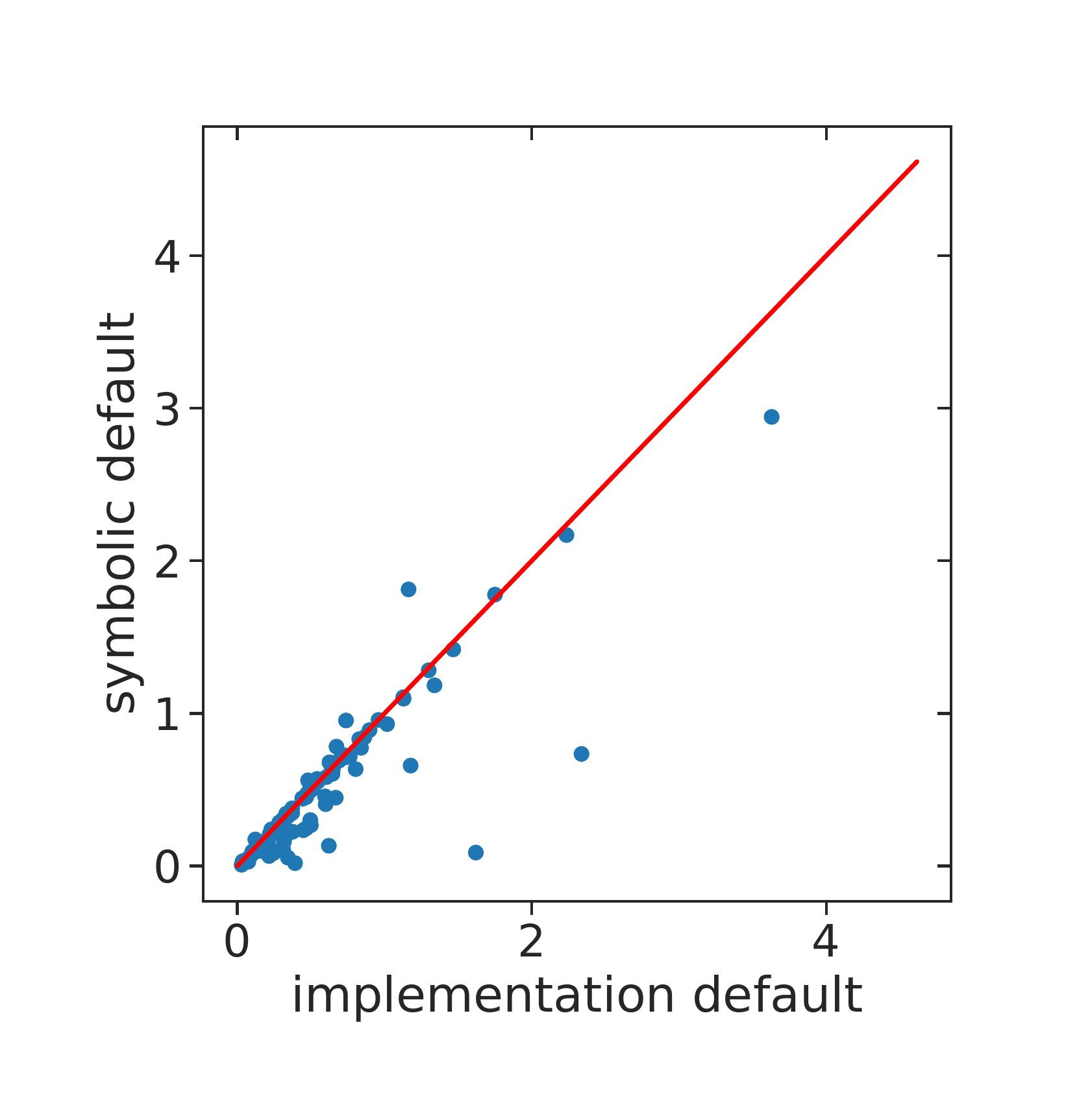}
\end{subfigure}
    \caption{Results for the Elastic Net algorithm.
    Comparison of symbolic and implementation default using log-loss across all datasets performed on real data. Box plots (right) and scatter plot (left)}
    \label{fig:svm_realdata}
\end{figure}

\end{document}